\documentclass[10pt,journal,compsoc]{IEEEtran}
\ifCLASSOPTIONcompsoc
  \usepackage[nocompress]{cite}
\else
  \usepackage{cite}
\fi
\ifCLASSINFOpdf
\else
\fi
\usepackage{hyperref}
\usepackage{url}
\usepackage{ amsmath}
\usepackage[super]{nth}
\usepackage[pdftex]{graphicx}
\usepackage{pdfpages}
\usepackage{amssymb}
\usepackage{algorithmic}
\usepackage[vlined,boxed,ruled]{algorithm2e}

\begin{document}
%
\title{Region Based Approximation for High Dimensional Bayesian Network Models}
%
%
%
%

\author{ Peng Lin, Martin Neil, Norman Fenton 
        
 }

%
%

\markboth{IEEE TRANSACTIONS ON PATTERN ANALYSIS AND MACHINE INTELLIGENCE/ submitted version}%
{}
%



\IEEEtitleabstractindextext{%
\begin{abstract}
Performing efficient inference on Bayesian Networks (BNs), with large numbers of densely connected variables is challenging. With exact inference methods, such as the Junction Tree algorithm, clustering complexity can grow exponentially with the number of nodes and so computation becomes intractable. This paper presents a general purpose approximate inference algorithm called Triplet Region Construction (TRC) that reduces the clustering complexity for factorized models from worst case exponential to polynomial. We employ graph factorization to reduce connection complexity and produce clusters of limited size. Unlike MCMC algorithms TRC is guaranteed to converge and we present experiments that show that TRC achieves accurate results when compared with exact solutions.   
\end{abstract}

}

\maketitle

\IEEEdisplaynontitleabstractindextext

%
\IEEEpeerreviewmaketitle

\IEEEraisesectionheading{\section{Motivation and Contribution}\label{sec:introduction}}

%
%
%
%
\IEEEPARstart{P}{erforming} efficient inference on Bayesian network (BN) models with a large number of variables that are also densely connected (high order dependent), is a major computational challenge. With exact methods, such as the Junction Tree (JT) algorithm \cite{barberBRML2012,Jensen:2007:BND1}, the complexity depends on the size of maximal cluster of the triangulated graph  \cite{koller,Jensen:2007:BND1}, and the maximal cluster size can grow exponentially with the number of nodes.\\
An important way to reduce this complexity is to apply  factorization algorithms, such as binary factorization  \cite{NeilCF12}, to convert the dense model into a factorized model resulting in reduced connection complexity. However, the resulting factorized BN's tree-width \footnote{Tree-width is one less than the minimum possible value of the maximum cluster membership size over all possible triangulations.} remains high, and cluster size in the factorized BN remains exponential as clustering will result in a product of all factors involved in the maximum cluster.  \\ 
As exact inference on high tree-width models is generally intractable, there are several techniques that use the bounded tree-width of JTs, such as \cite{Bach01thinjunction,ElidanandGould}. These so called thin-JTs ensure tractability by using an upper bound of the tree-width. However, the overall performance on accuracy is not guaranteed, except where child nodes are deterministically related to their parents or where the modeller can make context-specific independence assumptions \cite{Darwiche:2009:MRB:1534901}.\\
When exact methods cannot be performed efficiently on high tree-width models, sampling based methods, such as Markov Chain Monte Carlo (MCMC) \cite{barberBRML2012,hastings70}, are used, but MCMC solutions usually have to be tailored to the problem and convergence is not guaranteed.\\
Motivated by the success of using region based approximation for spin class/grid models \cite{YedidiaFW05}, we employ region based approximation to reduce the clustering complexity for high tree-width factorized models \cite{NeilCF12,PengLin}. We present a general inference algorithm called Triplet Region Construction (TRC) based on region belief propagation and show how it can perform robust inference on BNs. In doing so a number of existing well known challenges (region choice, convergence and accuracy) encountered when using region based approximation are addressed. Most significantly our methods provide an algorithm where the clustering and efficiency complexity for factorized models is reduced from worst case exponential to polynomial.  \\
The paper is structured as follows: \\
In section 2, we introduce necessary BN notation and background and explain why it is sufficient for our proposed algorithm to focus on complete BN models via a uniquely defined binary factorized model which has lower order dependences.\\
In section 3, we introduce region based approximation methods and explain why previous region based algorithms cannot deal with the factorized models defined in Section 2. \\
In section 4, we show how region based approximation can be used to develop our proposed TRC algorithm and reduce the clustering complexity. \\
In section 5, we present experiments involving sparse BN and high tree-width factorized BNs to show the accuracy and robustness of TRC. We also contrast these results with those obtained using MCMC and discuss the accuracy achieved.\\
Section 6 concludes the paper and discusses extensions of the TRC algorithm.\\
The paper provides four key novel contributions:      
\begin{enumerate}
	\item Whereas region based belief propagation \cite{YedidiaFW05} is typically applied to undirected graphical models here we present its use, in a systematic way, for directed models for the first time (sections 3 and 4).   
	\item  Construction of a region graph for general models is difficult because the choice of regions and interactions is left up to the model designer, with varied results. We present a region identification algorithm, called Outer Region Identification (ORI), that incorporates all local (considered conditional independence) factor correlations as an effective way of identifying the largest regions (sections 4.1 and 4.2). ORI can be used separately to provide the region specification for many other region graphs. We then use redundant regions resulting from ORI, to adjust our region graph to satisfy the perfect correlation property \footnote{This requires the sum of all regions' counting numbers to be one, i.e. $\sum\nolimits_R {{c_R} = 1} $ which ensures that the region-based entropy is correct if all variables in $p$ are perfectly correlated.} \cite{GelfandW12,YedidiaFW05} and maxent-normal property \footnote{A constrained region-based free energy approximation is maxent-normal if it is valid and the corresponding constrained region-based entropy ${H_{\cal G}}$ achieves its maximum when all the beliefs are uniform.} \cite{YedidiaFW05} that are necessary conditions for computational accuracy.
	\item Previous region based algorithms suffer from unavoidable numerical instability problems when performing inference on high tree-width factorized models. We propose a Region Graph Binary Factorization (RGBF) algorithm to decompose the region graph into an equivalent, but more numerically stable, alternative. We show that RGBF improves the robustness of region based belief propagation algorithms (section 4.4). RGBF is also a separate algorithm that can be used for any region graph. 
	\item Finally (and most importantly) we describe the TRC algorithm (section 4.5) in terms of the above sub-algorithms. TRC is guaranteed to converge, solves the complexity challenge encountered in clustering high tree-width factorized models and achieves accurate results when we compare the marginal distributions of individual variables with those produced by JT and with MCMC (section 5).  	
	
\end{enumerate}   

\section{Complete BN and Its decomposition} 
In this section we provide a brief overview of BNs and their notation and then discuss why it is sufficient to solve the BN inference problem using a complete BN graph, and how a BN's connection complexity can be reduced using binary factorization.\\
A BN is a directed acyclic graph (DAG), with nodes $X_1, X_2,...,X_n$ representing random variables (which can be discrete or continuous \footnote{All continuous variables can be assumed to be discretized statically or dynamically \cite{NeilTM07,koller,Koller+al:UAI99,KozlovK97,PengLin}.}), together with a conditional probability distribution (CPD) for each node which is conditional on its parent nodes if there are any (for discrete variables we refer to node probability tables (NPTs)). The absence of arcs between nodes encodes the Conditional Independence (CI) \cite{barberBRML2012} assumptions between variables. The BN represents the joint distribution, $p$ of the random variables $X_1, X_2,...,X_n$ as the product of its CPDs. In the absence of CI assumptions, we can use the chain rule to factorize the joint distribution, as shown in equation  (\ref{equa1}).
\begin{equation}
p(X_1,X_2,...,X_n)=\prod_{i=1}^{n}p(X_i|X_1,...,X_{i-1}) 
\label{equa1}
\end{equation}
With CI assumptions this simplifies to \[P({X_1},...,{X_n}) = \prod {P(} {X_i}|pa({X_i})\] where $pa\{X_i\}$ represents the parents of node $X_i$. This simplification, along with the associated graphical representation, is one of the attractions of using BNs.\\
However, in the worst case there are no CI assumptions in the BN. In this case the BN graph is a \emph{complete DAG} with $n$ nodes, i.e. every pair of nodes is connected by a directed edge. Performing inference on such a complete BN graph represents the worst case complexity for exact algorithms and is usually intractable. Since it is therefore assumed to be impossible to find efficient exact algorithms for arbitrary BN models, the challenge is to find good approximate algorithms.\\
It is crucial to note that any BN model can be regarded as a complete graph with some edges removed, where the remaining edges encode the CI assumptions. Conversely, any non-complete BN graph model (referred to as a sparse graph) can be converted to a complete graph model by adding appropriate edges. So, theoretically, any BN model can be represented by a complete graph.  Hence, any algorithm that performs efficient inference for complete BN models will also be efficient for arbitrary BN models. It is therefore sufficient to find an approximate inference algorithm that is efficient for complete BN models and hence, this is the focus for the rest of the paper.\\
In what follows we also make use of a well known result of graph theory (which can be proved by induction on the number of nodes) which asserts that any complete DAG of $n$ nodes has a unique Hamiltonian path, and is hence uniquely defined up to a permutation of the $n$ nodes. Specifically,\\ \\ 
\textbf{Theorem 1:} In any complete DAG of $n$ nodes there is exactly one node with indegree $n-1$, exactly one node with indegree $n-2,...,$ exactly one node with indegree one, and exactly one node with indegree zero. \\ \\
Theorem 1 ensures the uniqueness of the chain rule factorization (Equation 1) for a complete DAG of $n$ nodes subject to the order in which, for each $i=1,...,n$ node $X_i$ is the (unique) node with indegree $i-1$. In what follows we will assume this ordering of the nodes in the complete graph.\\
In addition to assuming a complete BN model we also need to transform the complete BN graph model into a version that is binary factorized (and equivalent in the sense defined below) by introducing additional nodes in such a way that each node has at most two parents. We call the process of producing a Binary Factorized BN (a BFG), the BF-process. A BFG avoids the computational complexity problem of exponential size CPDs (although, at this stage, the cluster size is not reduced). Instead each CPD has at most three members, i.e is at worst a triplet. Furthermore, we will also benefit from this complexity reduction when applying the region based approximation discussed in section 4. \\ \\ 
\textbf{Proposition 1:} A BN $G$ can be transformed into a binary factorized BN $G'$ (i.e. each node has at most two parents) whose nodes are a superset of $G$ and which is `equivalent' to $G$ in the sense that, for each node $X$ in $G$, the CPD of $X$ in $G'$ is equivalent after factorization to the CPD of $X$ in $G$. \\ \\
\emph{Proof}.  In what follows we assume the unique ordering of the complete graph G from Theorem 1, and apply the structural factorization of G as described in \cite{NeilCF12}, by introducing a set of intermediate variables $E_t$ (${E_t} \in G',{E_t} \notin G$) that are not in the original BN. For example, in the case of the 5-dimensional complete graph, the structure of the binary factorized version is as shown in Figure 1. While this BF algorithm is guaranteed to produce a uniquely structured BFG $G'$ or each complete BN graph $G$, we have to show how to define the CPDs in $G'$ so that for  each node $X$ in $G$, the CPD of $X$ in $G'$ is equivalent after factorization to the CPD of $X$ in $G$. There are three types of nodes whose CPDs we have to consider: 1) Continuous nodes with continuous parents only; 2) Discrete nodes with discrete parents only; 3) Mixture nodes (Continuous nodes with at least one discrete parent or discrete nodes with at least one continuous parent):   
\begin{enumerate}
\item \textbf{Continuous nodes with continuous parents:} The result for this case was proved in \cite{NeilCF12}. In summary, we assume a continuous CPD $P(Z|pa\{ Z\})$ is always expressed as an arithmetical expression over $Z$ and $pa\{ Z\} $, and this expression can always be parsed incrementally by smaller expressions which involve only two variables. By introducing the intermediate nodes, this naturally results in a BF process for the continuous case. The BF process in \cite{NeilCF12} was motivated to ensure the same equivalence property as is required here. Figure 1, provides an example applied to the 5-dimensional complete graph in the simple case where all nodes are continuous linear functions of their parents. In the resulting BFG $G'$ (with the exception of the root nodes $X_1$ and $X_2$), each node has exactly two parents.
\item \textbf{Discrete nodes with discrete parents:} In general a discrete node $D$ with three discrete parents $A$, $B$ and $C$ can be transformed into an equivalent binary factorised form by introducing an intermediate node $E$ (with parents $A$ and $B$) that has $n\times m$ states $e_{ij}$ ($i=1,...,n$ and $j=1,...,m$) where $A$ has $n$ states $a_1,...,a_n$ and $B$ has $m$ states $b_1,...,b_m$. The NPT for $E$ is defined as: \[P(E = {e_{ij}}|{a_k},{b_l}) = \left\{ \begin{array}{l}
1\;\;if\;k = i\;and\;l = j\\
0\;\;otherwise
\end{array} \right.\;\] The NPT for node $D$ in $G'$ (with parents $C$ and $E$) is defined as: \[P(D|{e_{ij}},{c_k}) = {P_G}(D|{a_i},{b_j},{c_k})\] Figure 2 shows the full solution for the 5-dimensional complete graph. The method is applied iteratively when there are more than three discrete parents.
\item \textbf{Mixture nodes:} Suppose $Z$ is a mixture node. We consider the two cases:\\
a) $Z$ is a continuous node with at least one discrete parent. If $Z$ has more than one discrete parent then we can apply the BF process described above for the discrete node parents, to ensure an equivalent factorization of those nodes such that $Z$ has just one discrete parent $Y$. So we can assume $Z$ has exactly one discrete parent $y$ and that the CPD for $Z$ is $P(Z|pa\{ Z\} ) = \sum  P(Y = {y_i}) \cdot f({X_i})$ ($Y,{X_i} \in pa\{ Z\} $), where $X_i$ are continuous variables. Then $Z$ can be binary factorized by incrementally combining the $X_i$ densities, $i>2$. The CPD for the intermediate variable $E_k$ can be defined as: \[P({E_k}|pa\{ {E_k}\} ) = \frac{{P(Y = {y_i}) \cdot {X_i} + P(Y = {y_j}) \cdot {X_j}}}{{P(Y = {y_i}) + P(Y = {y_j})}}\] (${X_i},{X_j} \in pa\{ {E_k}\} $). The CPD of $Z$ can be recovered by incrementally combining $E_k$ with another parent ${X_q}$ ($q \ne i \ne j$) by using the same formula as defined for $P({E_k}|pa\{ {E_k}\} )$. \\
b) $Z$ is a discrete node with at least one continuous parent. In this case we can apply the BF process for continuous nodes to factorize on its parent nodes and guarantee $Z$ has only two parents. An example of the mixture node case is shown in Figure 3. Specifically, $X_1$ is a discrete node and the other nodes are continuous; for example, $X_5$ is a mixture node with one discrete parent and three continuous parents. $\square$  
\end{enumerate} 
\begin{figure}[h]

\quad \	\includegraphics[scale=.56]{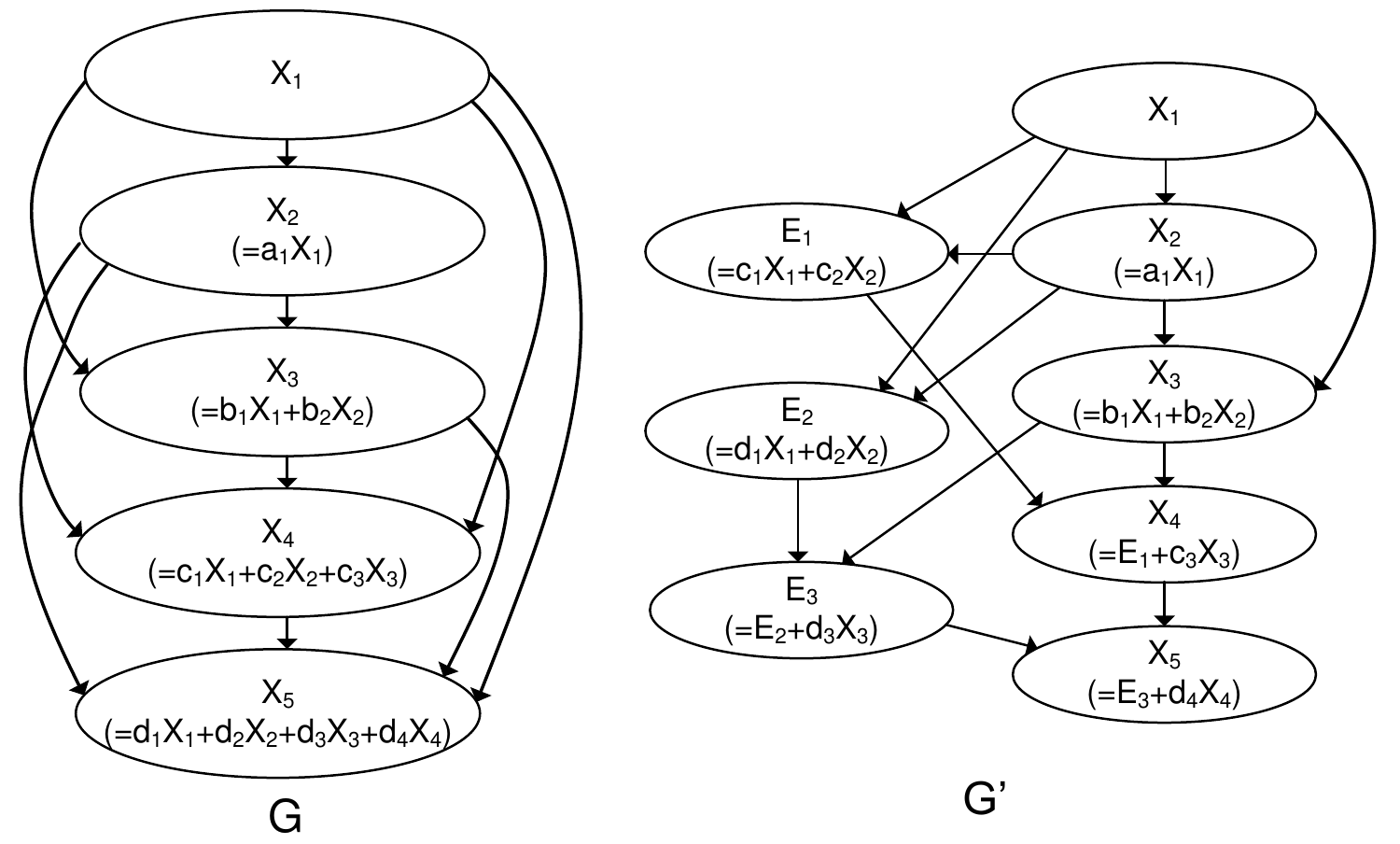}
 
\qquad\qquad\quad\quad	    (a) \qquad  \qquad\qquad\qquad \quad \qquad \ (b)
	\caption{(a) 5 dimensional complete graph $G$ with all continuous variables; (b) resulting BFG $G'$}
\end{figure}

\begin{figure}[h]
 
	\includegraphics[scale=.56]{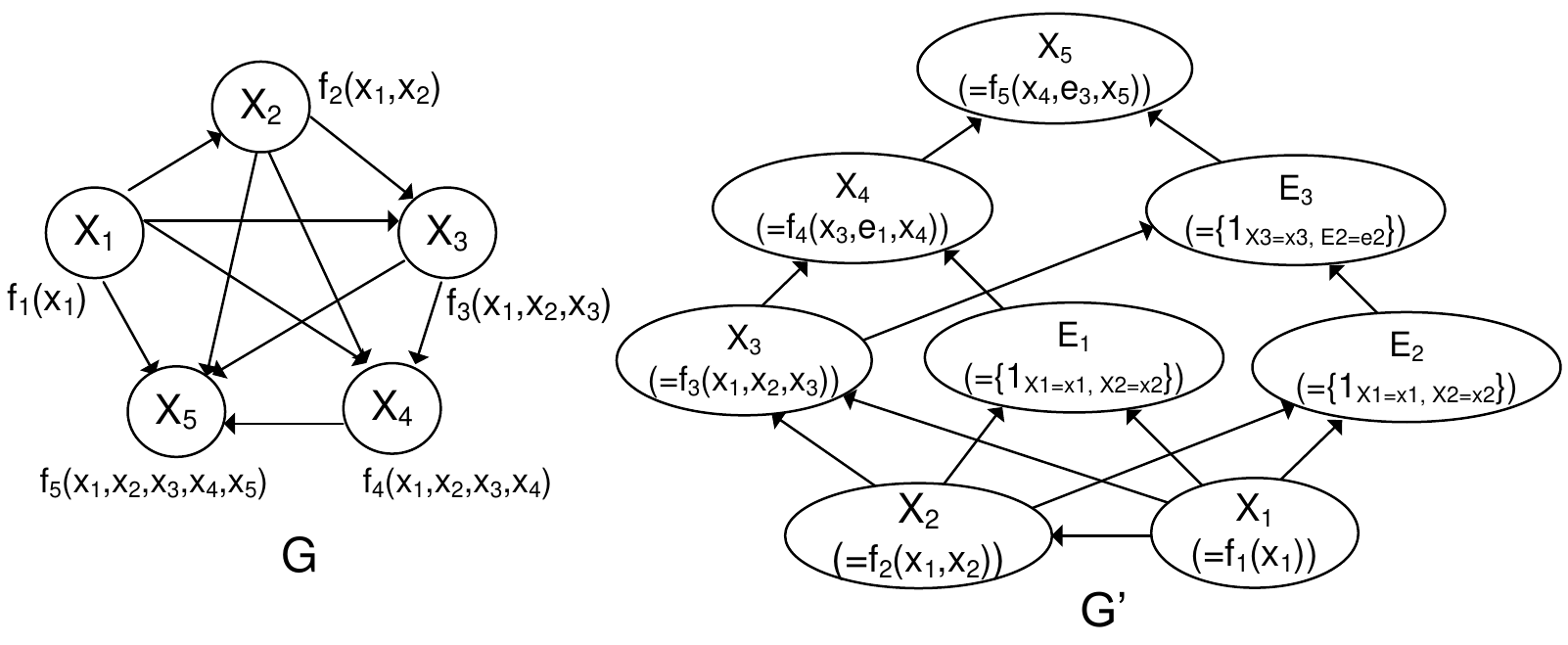}
	
\qquad  \qquad \ \ 	(a) \qquad  \qquad\qquad\qquad \qquad\qquad \ \  (b)
	\caption{(a) 5 dimensional complete graph $G$ with all discrete variables, nodes ${X_1},...,{X_5}$ are associated to CPDs ${f_1},...,{f_5}$; (b) resulting BFG $G'$}
\end{figure}
 \begin{figure}[h]
  
 	\includegraphics[scale=.56]{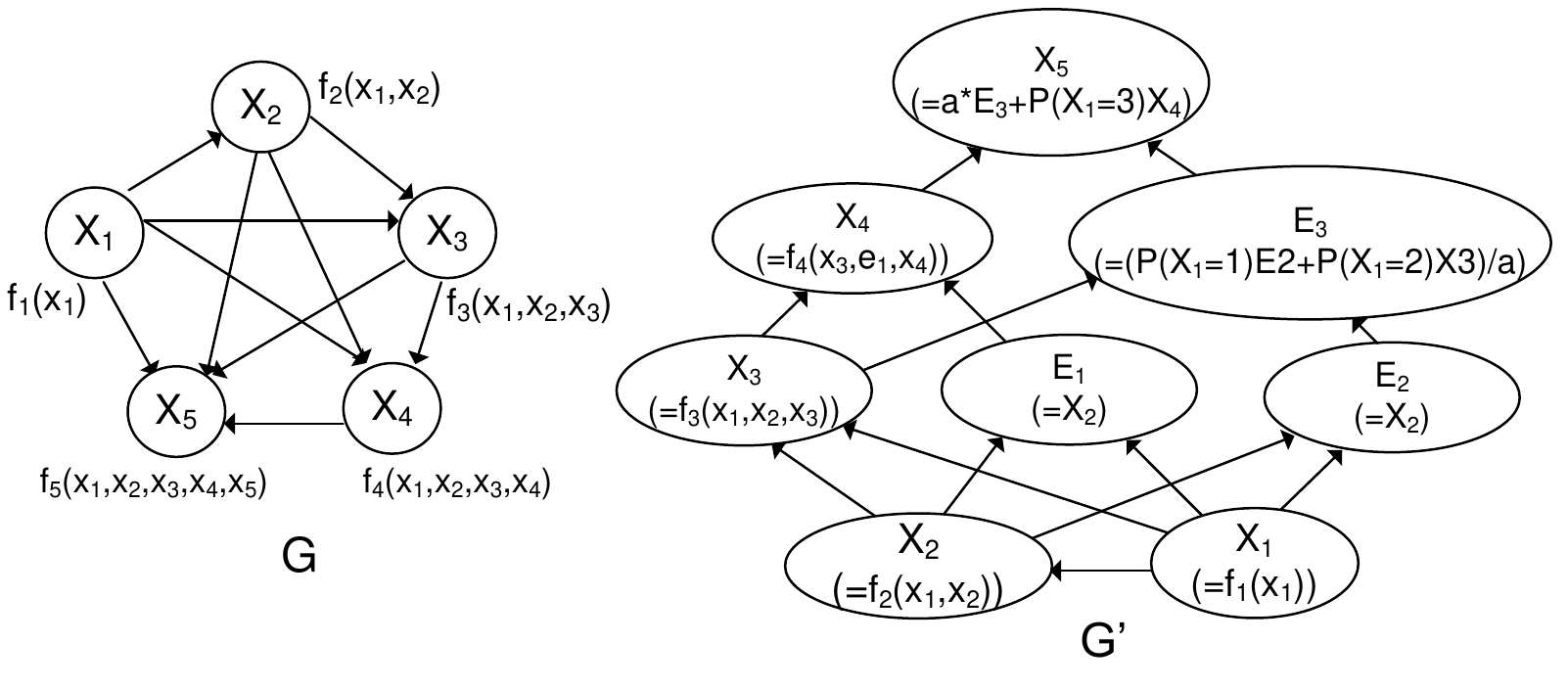}
 	
 \qquad  \qquad \ \	(a) \qquad  \qquad\qquad\qquad \qquad\qquad \ \  (b)
 	\caption{(a) 5 dimensional complete BN $G$ with nodes ${X_1},...,{X_5}$ and associated CPDs ${f_1},...,{f_5}$; (b) BF process for (a) if (a) is mixture with $X_1$ as the discrete variable, and others the continuous variables. The CPD for $X_5$ is defined as: ${f_5} = P({X_1} = 1) \cdot {X_2} + P({X_1} = 2) \cdot {X_3} + P({X_1} = 3) \cdot {X_4}$, with constant $a = P({X_1} = 1) + P({X_1} = 2)$. The function $f_4$ is not factorized but only has changed variables as only two continuous parent variables are involved. }
 \end{figure} 
 Applying the BF process to an $n$ node complete BN graph model results in a BFG model with ${\kappa _n}$ nodes where \[{\kappa _n} = n + (n - 2)(n - 3)/2 = ({n^2} - 3n + 6)/2\] We will use ${\kappa _n}$ to denote the number of variables in a BFG throughout the rest of the paper. 
\section{Region Based Belief Propagation}
As already noted, while binary factorization results in reduced factor size complexity and is a necessary step in the TRC algorithm that we propose, it does not solve the problem of cluster size complexity. Hence it does not avoid the complexity problems associated with exact algorithms, such as JT. When an n-dimensional complete BN is converted to a non-complete BN, the tree-width remains $n-1$ in the factorized BN. This is also evident in a $D \times D$ grid where the model is sparse but the tree-width is $O(D)$ \cite{Murphy:2012:MLP:2380985}. In these circumstances the triangulation graph produces clusters that are still (in the worst case) exponential.\\
In section 3.1 we will give an overview of region based approximation \cite{Minka:2001:FAA:935427,WainwrightJW03,YedidiaFW05,Darwiche:2009:MRB:1534901} which partly addresses the problem of cluster size complexity in our TRC algorithm, and section 3.2 discusses how it presents residual difficulties for high tree-width BN model problems (which we subsequently resolve in Section 4). 
\subsection{Review of GBP and CCCP Algorithms}
Before introducing region based belief propagation, we define region and region graph. \\\\
\textbf{Definition 1.} \textit{A region $ r $ of a factor graph is a set of variable nodes $V_r$ and factor nodes $A_r$, such that if a factor node $ a $ is in $ A_r $, all variable nodes neighbouring $a$ are in $A_r$.} \cite{YedidiaFW05}\\\\
\textbf{Definition 2.} \textit{A region graph $\mathcal{G}$ is a directed graph $\mathcal{G} = (\mathcal{V,\;E,\;L})$ in which each vertex $ v \in \mathcal{V} $ corresponding to a region $ r $ is labelled (denote the label of vertex $ v $ by $l(v) \in \mathcal{L}$) with a subset of nodes in a factor graph. We say ${v_p}$ is a parent of $ v_c $ if ${v_p} \to {v_c}$ is a directed edge $e \in \mathcal{E}$.} \cite{YedidiaFW05}\\\\
Region based belief propagation \cite{YedidiaFW05} is popularly used in specific undirected graphic models, such as finite dimensional spin class models \cite{Kolmogorov:2006:,Sun:2003:SMU,YedidiaFW05} encoded by a factor graph\footnote{A bipartite graph representing the factorization of a function, with factor node containing all factors.} These models may be intractable for exact methods but can be approximated using region based belief propagation. Here the clustering complexity reduction is achieved by constructing variational region based complexities and involves the approximation of a free energy function term and its function space.\\
Performing exact inference is equivalent to solving an optimization problem over the exact energy function $F[p,Q]$, where $p$ is a distribution over $\chi $ (the set of all possible assignments of values to all the network's random variables) and $Q$ is the space of all marginals. This is NP-hard \cite{Darwiche:2009:MRB:1534901,Kolmogorov:2006:}. When $F[p,Q]$ cannot be tractably optimized a factored energy function can be defined in terms of entropies over all regions in a region graph ${\cal G}$, as an approximation of $F[p,Q]$. However, even approximation of $F[p,Q]$ over the marginal polytope $marg[{\cal G}]$ ($marg[{\cal G}]=\{Q_p\}$) is also an NP-hard problem\cite{Darwiche:2009:MRB:1534901}. Instead, we perform optimization over the locally consistent polytope $local[{\cal G}]$, which is a set of pseudo-marginal distributions over the variables in each region. This local consistency achieves a polynomial computation complexity provided that all regions are calibrated and neighbouring regions are locally consistent with each other \cite{Darwiche:2009:MRB:1534901}.\\
Yedidia et al. \cite{YedidiaFW05} demonstrated that the convergence  of these self-consistent constrained region belief equations corresponds to the minimal points of the Kikuchi variation free energy. Minimization of the Kikuchi cluster free energy is equivalent to the problem of constructing a fixed point for the region belief equations and this can be achieved using iterative message-passing over a region graph. In this section, we review two of the region based message-passing algorithms \cite{MateescuKGD10,Minka:2001:FAA:935427,YedidiaFW05,Yuille02cccpalgorithms} that we use as the starting point for our approach. The first is Generalized Belief Propagation (GBP) \cite{YedidiaFW05}, which is a generalization of a class of belief propagation based algorithms, involving Loopy Belief Propagation \cite{Murphy:1999:LBP}, Survey Propagation \cite{Braunstein:2005}, and others. However, GBP does not guarantee convergence \footnote{It is possible to use the so-called dumping technique \cite{YedidiaFW05,Jaimovich:2010} to help converge but it is still not guaranteed.}, and although the second message passing algorithm, called the Concave Convex Procedure (CCCP) \cite{Yuille02cccpalgorithms,Yuille:2003:CP:773700.773708}, does guarantee convergence it can, unfortunately, be numerically unstable for large models. Indeed, performing inference on the region graph for our high tree-width BFG models using both algorithms is numerically unstable and/or does not converge. Other BP related algorithms, such as expectation propagation \cite{Minka:2001:FAA:935427}, non-parametric belief propagation \cite{Sudderth:2010:NBP}, particle belief propagation \cite{nips09}, are not discussed here as they characterize messages in a continuous domain only with additional approximations and also do not guarantee convergence. \\
For convenience to use GBP we can apply the Cluster Variation Method (CVM) \cite{YedidiaFW05} to produce a valid region graph. Firstly the outer regions \footnote{An outer region is defined to be those regions with no parents, i.e. having no incoming region edges.} are identified for the first level of a region graph and next CVM generates the regions for subsequent levels using intersections of the regions declared at the previous level. The Kikuchi region based free energy function $F_\mathcal{G}$ is defined in Equation 2. Our task is to minimize $F_\mathcal{G}$ under a set of self-consistent constraints imposed by the CVM region graph.
\begin{align} 
{F_\mathcal{G}} = &\sum\limits_{r \in R} {c_r} \{ \sum\limits_{{x_r}} {b_r}({x_r}){E_r}({x_r}) + \nonumber   \\
&\sum\limits_{{x_r}} {{b_r}({x_r})\log {b_r}({x_r})}  \}   + {L_\mathcal{G}} \label{equa2}
\end{align}
where $R$ is the set of all regions in ${\cal G}$, ${c_r} = 1 - \sum\limits_{r' \in Ancestor(r)} {{c_{r'}}} $, and corresponds to counting number of each region, given ${c_{r'}}$ is number of degrees of freedom for the region. The term ${E_r}({\bf{x}})$ represents the energy associated with region $r$. The region belief term $b_r$ is an estimated distribution of the true distribution over region $r$ \cite{YedidiaFW05}.\\\\
To perform minimization of Equation 2 we need to consider the Lagrangian term ${L_{\cal G}}$ (shown in equation 3), which incorporates two kinds of constraints: the normalization constraint for each region, $\sum\limits_{{x_r}}  {b_r}({x_r}) = 1$ and the running intersection constraints between parent and child region beliefs, $\sum\limits_{x \in r\backslash c}  {b_r}({x_r}) = {b_c}({x_c})$ ($c \in child(r)$).
\begin{align}  
{L_\mathcal{G}} =&  \sum\limits_{r\in R} \sum\limits_{c \in child(r)}\sum\limits_{{x_c}} {{\lambda _{r,c}}}({x_c})\{\sum\limits_{x \in r\backslash c} {{b_r}({x_r}) - {b_c}({x_c})}\}  \nonumber  \\
&+\sum\limits_{r \in R}{\gamma _r}(\sum\limits_{{x_r}}{{b_r}({x_r})- 1}) \label{equa3}
\end{align}
Solving for Lagrangian multipliers $\lambda$ and $\gamma$, in (3), corresponds to an iterative message passing algorithm. Both GBP and CCCP updated equations involve solving Lagrangian multipliers. The difference between CCCP and GBP is that CCCP splits the free energy function $F_\mathcal{G}$ into a concave and a convex part, ${F_{\cal G}} = {F_{{\cal G},cave}} + {F_{{\cal G},vex}}$ and GBP does not. Minimization of the free energy function is then an iterative procedure guaranteed to minimize the convex part and maximize the concave part using tangent matching \cite{Yuille02cccpalgorithms}. In general, GBP is more efficient than CCCP since CCCP uses an outer-inner double loop and each inner loop involves recursively updating the Lagrangian multipliers.
\subsection{Region Based Approximation Difficulties for Directed Models}  
Region graphs generated for high tree-width directed models often involve multiple connections between regions located at different levels. Because one parent can have a large number of children the same variable will appear in many different regions across these levels. This gives rise to a large counting number and multiple cycles associated with a single region, leading to under/overflows during the multiplication of multiple messages. Such numerical instability might be encountered during message updating in both GBP and CCCP. The message updating process is shown in equations 4, 5 and 6 for GBP.   
 \begin{align} 
 {b_r}({x_r}) = {\tilde f_r}({x_r})\prod\limits_{c \in child(r)}  {n_{c \to r}}({x_c})\prod\limits_{p \in parent(r)}  {m_{p \to r}}({x_p})\label{equa4}
 \end{align}
 \begin{align}
 {m_{p \to r}}({x_r}) = (n_{r \to p}^0({x_r}{))^{{\beta _r} - 1}}{(m_{p \to r}^0({x_r}))^{{\beta _r}}}
 \label{equa5}
 \end{align} 
 \begin{align}
 \begin{array}{*{20}{l}}
 {m_{r \to c}^0({x_c}) = }&{\sum\limits_{{x_r}\backslash {x_c}}  {{\tilde f}_r}({x_r})\prod\limits_{p \in parent(r)}  {m_{p \to r}}({x_r})}\\
 {}&{\prod\limits_{c' \in child(r)\backslash c}  {n_{c' \to r}}({x_{c'}})}
 \end{array} \label{equa6}
 \end{align}  
where ${\tilde f_r}({{\bf{x}}_r}) \equiv {(\prod\nolimits_{a \in {A_r}} {{f_a}} ({{\bf{x}}_a}))^{{c_r}}}$, ${\beta _r} = 1/(2 - (1 - {c_r})/{p_r})$ (${p_r}$ is the number of parents of $r$) and ${n_{c \to r}}$ is similarly updated as $m_{p \to r}$ by pseudo-messages ($m_{r \to c}^0({x_c})$).  \\\\
As all messages are exponential family distributions, a large absolute value of counting number $|c_r|$ will result in either large or small values of ${\tilde f_r}$, which in turn influences $m{}_{p \to r}$ and also $b_r$. The over/under flow problem is unavoidable when the number of dimensions is large. Likewise, a large counting number is caused by  multiple paths from parents to the same child, resulting in many cycles in the region graph. Too many cycles in a region graph will make message scheduling difficult and inhibit convergence. The region graphs constructed for high tree-width BFG models inevitably encounter these large counting numbers and multiple cycles problems, and as a result both GBP and CCCP are indeed numerically unstable.  \\
In summary, there are three major difficulties preventing the use of region based belief propagation to approximate high tree-width BFG models:
\begin{enumerate}
	\item The construction of region graphs has, to date, been ad-hoc and problem specific, thus making generalisation and accuracy difficult. We address this in sections 4.1 and 4.2. 
	\item It is not clear on how much interaction strength\footnote{This is the number of interactions between outer regions.}\cite{Welling04} among outer regions is needed. This problem is also addressed in sections 4.1 and 4.2.
	\item By using CVM to generate regions for high tree-width BFG models the same variables can appear in multiple regions in the region graph, leading to multiple cycles associated with the smallest regions  and numerical instability. This problem is addressed in section 4.4.
\end{enumerate}

\section{Triplet Region Construction} 
This section summarises the TRC algorithm, designed to address the difficulties that arise when using region based approximation for high tree-width directed models. TRC is composed of three sub algorithms. In Section 4.1, we provide a formulation for converting a BN to a parametric Markov Network and we propose the Outer Region Identification Algorithm (ORI) to identify outer regions as a first step to construct a valid region graph. In Section 4.2, we illustrate how redundant outer regions are identified and can be rejected, so we can produce a region graph which will satisfy desired properties for accuracy. Section 4.3 proves that the region graph we proposed satisfies the desired properties for accuracy. To avoid numerical instability, in Section 4.4 we propose the Region Graph Binary Factorization (RGBF) algorithm to ensure each region has exactly two parent regions. Section 4.5 describes the TRC algorithm as a combination of ORI, RGBF and CCCP. 
\subsection{Outer Region Identification Algorithm}
Since belief propagation algorithms are typically designed for undirected models, such as Markov Networks\footnote{A Markov Network is a set of random variables having a Markov property described by an undirected graph.} (MNs), for directed models we need to convert the CPDs into factors. \\
To convert a BN to an undirected parameterization we first identify factors $\phi$ such that ${\phi _{\{ {X_i}\}  \cup pa\{ {X_i}\} }}({X_i},pa\{ {X_i}\} ) = P({X_i}|pa\{ {X_i}\} )$. \\
We also need to construct the \textit{moral graph} of a BN $G$, denoted $M[G]$. This is an undirected graph that contains an edge $(X_i, X_j)$ if there is an edge between $X_i$ and $X_j$ in $G$, or if $X_i$ and $X_j$ are parents of the same child node\footnote{Because a factor in undirected models is defined on all variables it contains.}. The added edge between the parents that share the same child node is called a \textit{moral edge}. \\ 
Unfortunately, connecting the parent nodes $X_i$ and $X_j$ via a moral edge assumes that $X_i \not\perp X_j$ and we may lose CI information contained in the original BN (i.e. $I(M[G]) \subseteq I(G) $). To resolve this, the CI information will be incorporated during the region graph construction.\\
Constructing a good region graph is an open research question, because minimizing the free energy function is a necessary but not sufficient condition. For instance, Welling et al.  \cite{GelfandW12,Welling04,WellingMT05} discuss ways to produce Structure Region Graphs based on graphical topology and they offer guidance based on structural information criteria. They define outer regions based on basic cycles in the MN graph. Interaction is optimal when the cycles can be ordered so that each cycle has some edge that does not appear in any cycle preceding it in the ordering. However, optimization from graphical topological information is difficult in the absence of some general rules.\\
Therefore, the choice of outer regions is a vital first step and if this is incorrect or suboptimal convergence and stability cannot be guaranteed.\\
We call our approach to identifying the outer regions \textbf{Outer Region Identification (ORI)}, which identifies outer regions in terms of maximizing interaction strength as well as reducing redundancy. 
\\
As $p'$ is already represented by the product of only triplet factors derived by the BF algorithm, any factors larger than triplets can be decomposed into triplet factors. Due to the Markov property this is an exact procedure for the decomposition. As a result, we can include all triplet factors as outer regions at the first level of a CVM region graph, which will produce a valid region graph \cite{YedidiaFW05}. The remaining problem is to determine the interaction strength among these outer regions. For this we need to introduce the \textit{Maximally Exhaustive} property.   \\ \\
 \textbf{Definition 3.} \textit{A region graph satisfies the Maximally Exhaustive property if any maximum subset of the outer region that contains a factor converted from the original BN is included in at least one second level region.}\\ \\
 \textbf{Theorem 2:} In a valid CVM region graph for BFG models with all triplet factors as outer regions interaction strength is sufficient if the second level regions are maximally exhaustive subsets of the outer regions. \\ \\
 \textit{Proof.} Because the maximum membership subset of a triplet factor is a node pair, the maximum interaction between any two triplet outer regions is a pair-wise interaction. The maximally exhaustive property ensures the number of local pair-wise interactions is at the maximum. Thus, interaction strength among all outer regions is sufficient in the sense of the number of interactions.$\square$\\\\ 
 To satisfy Theorem 2 we first need to generate local pair-wise interactions, which is achieved by defining two types of outer region members for our BFG models: primary triplets and interaction triplets. \\\\
 \textbf{Definition 4.} \textit{A {Primary Triplet} ${\cal F} = ({{\cal V}_{{X_i}}},\phi )$ is a triplet with nodes set ${{\cal V}_{{X_i}}} = \{ {X_i},{X_j},{X_p}\} $ in the moral graph $M[G']$ and a factor $\phi $ defined by the conversion from the CPD $P({X_i}|{X_j},{X_p})$ in the BFG, $ G' $, as a child variable $X_i$ and its two parents $X_j$ and $X_p$.}\\ \\
 \textbf{Definition 5.} \textit{An {Interaction Triplet} $\mathcal{U} = (\mathcal{V},\phi )$ is a triplet with factor $ \phi $ defined as a uniformly distributed factor, and triplet nodes ${\cal V}  \in M[G']$ where ${\cal V}$ is not equal to any primary triplet's nodes set.}\\ \\
 We also need to define:\\
 \textbf{Definition 6.} \textit{A Maximum Membership Subset of a primary triplet ${\cal F} = ({{\cal V}_{{X_i}}},\phi )$, ${\Omega _{{X_i}}}$ ($X_i$ is child node of $X_j$ and $X_p$), is the set of combinations of all node pairs in ${{\cal V}_{{X_i}}}$: $\{ {X_i},{X_j}\} $, $\{ {X_i},{X_p}\} $ and $\{ {X_j},{X_p}\} $.}\\\\
 The shared nodes between two primary triplets will mostly contain a single node only and this obviously fails to satisfy Theorem 2. Instead, by adding interaction triplets we can use node pairs that belong to different primary triplets, which creates a maximum membership subset via which two or more primary triplets can interact. This method of adding new regions to create interactions is also evident in \cite{Welling04}.
 \begin{figure}[h]

 \quad	\includegraphics[scale=.47]{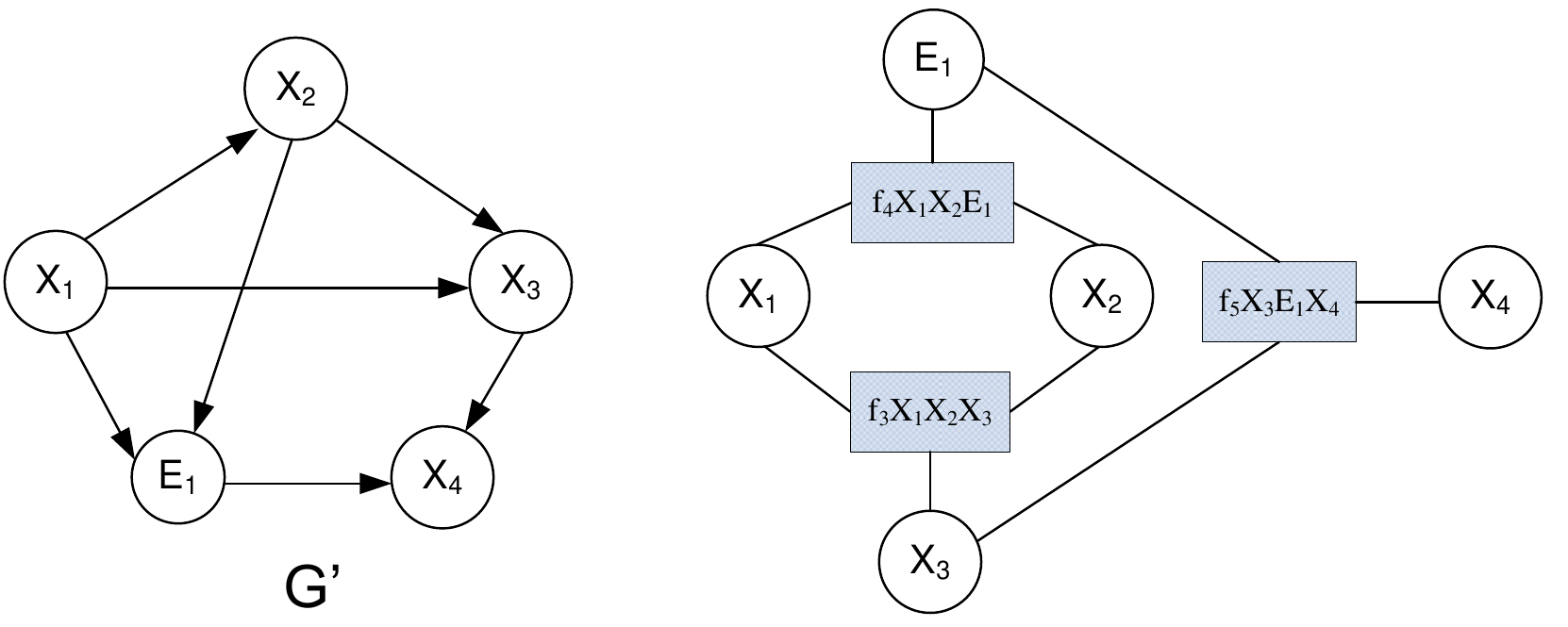} 
 	
 \qquad\qquad \quad  (a) \quad\qquad\quad \qquad \  \qquad (b)  
 	\caption{(a) ${\kappa _4}$ BFG; (b) factor graph of (a), with factor ${f_1} = {\phi _{{X_1}}}$ and ${f_2} = {\phi _{{X_1}{X_2}}}$ multiplied into $f_3$   }
 \end{figure}
\\Figure 4 shows an example where pair-wise interaction is identified using only primary triplets factors, but does not meet the maximally exhaustive property. The factor graph shows a cycle propagation on $X_1$ and $X_2$. This means $X_1$ and $X_2$ are strong pair-wise correlated and can be identified as a pair-wise interactions connecting $\{ {X_1}{X_2}{E_1}\} $ and $\{ {X_1}{X_2}{X_3}\} $ primary triplet regions. However, it is not explicitly known to what extent the node pair $\{ {X_2},{E_1}\} $ (a maximum subset of $\{ {X_1}{X_2}{E_1}\} $) is also pair-wise correlated, since $\{ {X_2},{E_1}\} $ is not cyclicly propagated between factors and is not shared by any two primary triplets. This  problem of lack of interaction can be fixed by adding an interaction triplet region $\{ {X_2}{X_3}{E_1}\} $. Similarly, other pair-wise correlations will also be incorporated by adding other interaction triplets. \\
Hence, second level regions in our region graph will be an exhaustive set of all possible pair-wise interactions among all triplet factors. \\
Although Theorem 2 provides a sufficient condition for the number of interactions, there are also other necessary conditions required for the accuracy of a region graph, as Theorem 3 shows.\\\\
\textbf{Theorem 3}. A region graph that does not satisfy both the perfect correlation property and maxent-entropy normal property will not be computationally accurate \cite{YedidiaFW05}. \\\\
Theorem 3 has been informally proved in \cite{YedidiaFW05} and is usually used as the necessary conditions to guide  region choice.   \\
Next, our ORI algorithm will first identify all outer regions that satisfies Theorem 2 and then identify some redundancy regions to reject, in order to satisfy Theorem 3. 
 As all primary triplets factors are CPD conversions from a BFG, they are already identified in the moral graph. All we need to do next is to identify the interaction triplets and this is achieved by using a coupled Markov Blanket\footnote{Markov Blanket for a node $X$ is a set of BN nodes that is composed of its parents, its children and its children's other parents, to guarantee conditional independence between nodes inside and outside of the set \cite{koller}.} \\ \\
 \textbf{Definition 7.} \textit{A coupled Markov Blanket for nodes $({X_i},{X_j})$ is the set of nodes $\partial ({X_i},{X_j})$ composed of ${X_i}$ and ${X_j}$'s Markov blanket excluding nodes $({X_i},{X_j})$. Therefore $\partial ({X_i},{X_j}) = \partial {X_i} \cup \partial {X_j}$ and $\partial ({X_i},{X_j}) \cap ({X_i},{X_j}) = \emptyset $.
 }\\\\
 The coupled Markov Blanket limited the number of candidate nodes that will be used to generate candidate interaction triplets for a node pair $\{ {X_i},{X_j}\} $. The Markov property encoded by a coupled Markov Blanket ensures that our candidate interaction triplet is optimally localized (considered CI information) to capture the local pair-wise correlations.
 \begin{algorithm}[!htbp]
 	\SetCommentSty{small}
 	\caption{ORI algorithm}
 	\KwIn{$p'$ factorized by $G'$,\ $p'(x) = \prod\limits_{i \in V}  P({X_i}|pa\{ {X_i}\} )$ \ $=\prod\limits_{i \in V} {\phi _{\{ {X_i}\} \cup pa\{ {X_i}\} }}({X_i},pa\{ {X_i}\})$}
 	\KwOut{Interaction triplets ${\cal U}$}
 	\textbf{Initial:} \ Interaction triplet ${\cal U} \leftarrow \emptyset $\;
 	\qquad\quad\quad all primary triplets ${{\cal F}_i} = ({{\cal V}_{X_i}},{\phi _i})$ by\;
 	\qquad\quad\quad  ${{\cal V}_{X_i}} \leftarrow \{ {X_i}\}  \cup pa\{ {X_i}\} $\; 
 	\quad\quad\quad \ \ \ \ ${\phi _i} \leftarrow {\phi _{\{ {X_i}\}  \cup pa\{ {X_i}\} }}({X_i},pa\{ {X_i}\} )$\;
 	
 	\For{each ${{\cal F}_i}$, the maximum subsets ${\Omega _{{X_i}}} \subset {{\cal V}_{X_i}} $}
 	{
 		\For {each node pair $({X_a},{X_b}) \in {\Omega _{{X_i}}}$ }{
 			
 			\For{each node ${X_c} \in \partial ({X_a},{X_b})$}{
 				
 				${\cal U} \leftarrow {\cal U} \cup \{ ({X_a},{X_b}) \cup {X_c}\} $ $(a \ne b \ne c)$\;
 			}
 						
 		}
 		
 	}
 
 	\For{each interaction triplet ${{\cal U}_i} \in {\cal U}$}{
 		\If{${{\cal U}_i}$ contains a node pair that is not directly connected in $M[G']$}{
 			
 			Reject ${{\cal U}_i}$;\
 			}
 	 
 			\If{${{\cal U}_i}$ contains a node pair that is a moral edge in $M[G']$ and ${{\cal U}_i}$  contains root node of $G'$ }{
 				
 				Reject ${{\cal U}_i}$;\
 			}
 		
 		}
 	return ${\cal U}$\;
 \end{algorithm}
 \\ 
In Algorithm 1 (ORI) each node pair $\{ {X_a},{X_b}\}  \in {\Omega _{{X_i}}}$ has a coupled Markov Blanket $\partial ({X_a},{X_b})$. Interaction triplets $\{ {X_a},{X_b},{X_c}\} $ are then identified for each node pair $\{ {X_a},{X_b}\} $ and each node $X_c$ in $\partial ({X_a},{X_b})$, ${X_c} \in \partial ({X_a},{X_b})$ $(a \ne b \ne c)$.\\
 ORI ensures all local pair-wise correlations are incorporated by introducing interaction triplets to exhaust all possible local pair-wise interactions (including a node pair that is connected as a moral edge in $M[G']$). So the conditional dependence information, imposed during conversion of the BN to the corresponding MN, is incorporated. Each interaction triplet is initialized with uniformly distributed factor.   
 \begin{figure}[h]
 
 	\centering
 	\includegraphics[scale=.45]{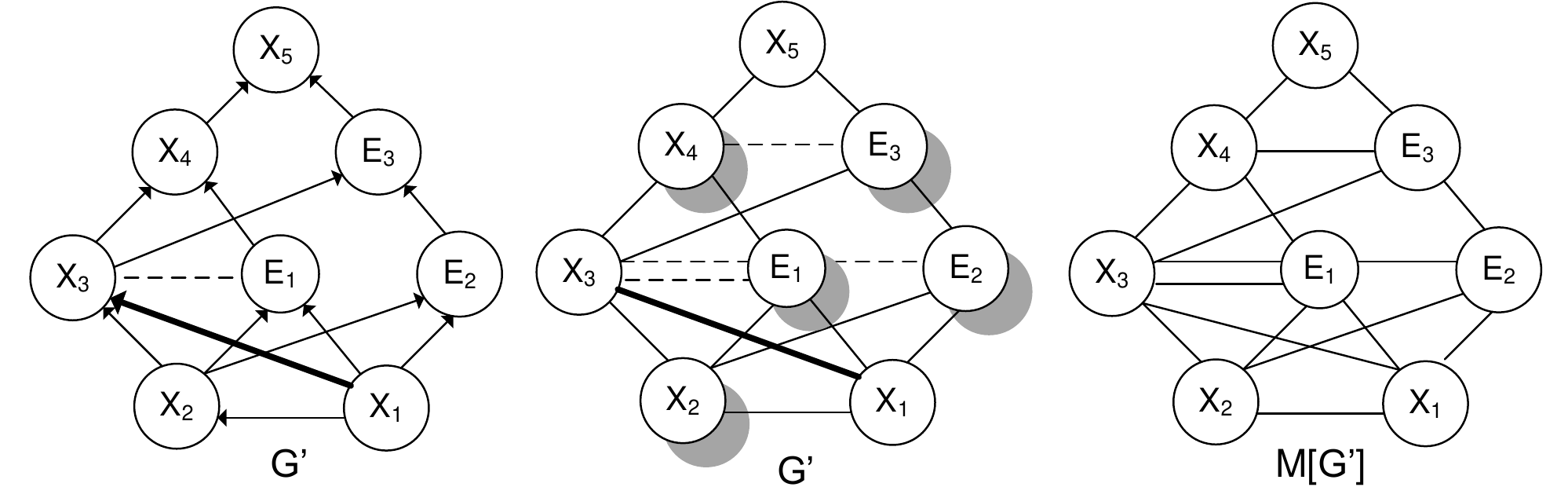}
 	\vspace{.001in}
  \quad \ \ (a) \qquad\qquad\qquad \quad  \ (b) \ \ \qquad\qquad\qquad\quad (c)
 	\caption{(a) ${\kappa _5}$ BFG with a moral edge shown as dashed line; (b) ${\kappa _5}$ BFG with directions removed and all moral edges shown as dashed lines; (c) moral graph of a ${\kappa _5}$ BFG }
 \end{figure}
 \\Figure 5 (a) shows a primary triplet $\{ {X_1}{X_2}{X_3}\} $ and an interaction triplet $\{ {X_1}{X_3}{E_1}\} $ interacted via an edge $({X_1},{X_3})$.\\
  In Figure 5 (b), for the primary triplet $\{{X_1}{X_2}{X_3}\}$, all its maximum subsets (or corresponding edges), $({X_1},{X_2})$, $({X_1},{X_3})$ and $({X_2},{X_3})$, which interact via primary triplet  ${X_1}{X_2}{X_3}$ and its interaction triplets. For example, to identify interaction triplets for edge $({X_1},{X_3})$ (shown as a bold solid line), we first identify the coupled Markov Blanket $\partial ({X_1},{X_3}) = \{ {X_2},{E_1},{E_2},{X_4},{E_3}\} $ (shown with nodes shadowed). From this the candidate interaction triplets are easily identified as: $\{{X_1}{X_3}{E_1}\}$, $\{{X_1}{X_3}{E_2}\}$, $\{{X_1}{X_3}{X_4}\}$, $\{{X_1}{X_3}{E_3}\}$ and $\{{X_1}{X_3}{X_2}\}$. Notice that $\{{X_1}{X_3}{X_2}\}$ is then excluded as it is a primary triplet. Repeated interaction triplets resulting from the selection of other edges' candidate interaction triplets will be removed.\\
   All primary triplets can be explicitly identified in the moral graph $M[G']$, and all maximum subsets, as node pairs, are identified as the edges in $M[G']$ as shown in Figure 5 (c).\\
   However, by selecting interaction triplets within a coupled Markov blanket we also introduced redundant triplets, which we will subsequently identify and remove. 
\subsection{Redundancy of interaction triplets}   
In Theorem 3 we introduced the perfect correlation property which requires that the sum over all counting numbers should be one. We identify two kinds of redundant interaction triplet regions that need to be rejected during outer region identification to satisfy this perfect correlation property:\\
Type 1, An interaction triplet that contains a node pair (edge) that does not exist in $M[G']$;\\
Type 2, An interaction triplet that contains a root node of $G'$ and a moral edge in $M[G']$.\\\\
Both types can be removed from the collection of all interaction triplets (as will be described in Proposition 3 below). \\
Because an interaction triplet $\{ {X_i},{X_j},{X_k}\} $ is defined to connect two or more primary triplets via their maximum subset (node pairs) the redundancy of an interaction triplet can be  determined by testing if the entropy of any variable in $\{ {X_i},{X_j},{X_k}\} $ changes after introducing interaction triplet $\{ {X_i},{X_j},{X_k}\} $, given other conditions are fixed.\\
To consider the entropy for each variable in $\{ {X_i},{X_j},{X_k}\} $ we first need to determine the relationship among the three node pairs, in order to analyse the relationship among variables $X_i$, $X_j$ and $X_k$. \\
The relationship among node pairs can be quantified by the mutual information\footnote{Region entropy is defined as $H({b_r}) \equiv  - \sum\limits_{_i} {{b_r}({x_i})\ln {b_r}({x_i})} $, mutual information for two region beliefs are $I({b_r};{b_s}) = H({b_s}) - H({b_s}|{b_r})$.} of node pair region beliefs. There are three node pairs in $\{ {X_i},{X_j},{X_k}\} $ and there is at least one node pair that is equivalent to a corresponding moral edge or that is not directly connected as an edge in $M[G']$. This means such a node pair $\{ {X_j},{X_k}\} $ has a pair-wise factor ${\phi _{j,k}} = 1$. This uniform factor encodes only conditional independence information $j \bot k|pa\{ j,k\} $ and hence $X_j$ and $X_k$ can be separately considered via other node pairings. So the mutual information between this node pair region to another node pair region is equal to the entropy over the shared variable. So the entropy over $X_j$ and $X_k$ will be determined by other two node pairs ($\{ {X_i},{X_j}\} $, $\{ {X_i},{X_k}\} $) after introducing $\{ {X_i},{X_j},{X_k}\} $, given other conditions are fixed.\\
Next we only need to consider the mutual information of the other two node pairs ($\{ {X_i},{X_j}\} $, $\{ {X_i},{X_k}\} $) which have non-uniform factors. Suppose ${b_{i,j}}$ and ${b_{i,k}}$ are region beliefs associated with region $\{ {X_i},{X_j}\} $ and $\{ {X_i},{X_k}\} $, given $j \bot k|pa\{ j,k\} $ we have:
\[\begin{array}{ll}
I({b_{i,j}};{b_{i,k}}) &= H({b_{i,k}}) - H({b_{i,k}}|{b_{i,j}})\\
&= H({b_{i,k}}) - H(\sum\limits_i {{b_{i,k}}} ) \\
&= H({{\tilde b}_i})
\end{array}\]
where $H({\tilde b_i})$ is the entropy of the marginal belief over variable $i$ in these two regions containing the node pairs. \\\\
In summary, $H({\tilde b_i})$ is mutual information that in turn determines the entropies of the two node pair regions. So we can reject an interaction triplet $\{ {X_i},{X_j},{X_k}\}$ by determining if $H({\tilde b_i})$ is changed by introducing $\{ {X_i},{X_j},{X_k}\} $ given other conditions are fixed.\\
Based on the above derivation over $\{ {X_i},{X_j}\} $ and $\{ {X_i},{X_k}\} $ we can prove:\\\\
\textbf{Proposition 3}. Type 1 and Type 2 triplets are redundant. \\\\
\textit{Proof}. We have an interaction triplet $\{ {X_i},{X_j},{X_k}\} $ connecting primary triplets $\{ {X_i},{X_j},{X_p}\} $ and $\{ {X_i},{X_k},{X_q}\} $ to determine $H({\tilde b_i})$ (${X_i}$ is short for $i$ in the following), where $j \ne k$. \\
We first show Type 1 redundancy which occurs only when $p \ne q$: \\
Given $p \ne q$, $\{ i,j\}  \cap \{ i,k\}  = \{ i,j,p\}  \cap \{ i,k,q\}=\{i\} $, the shared subset of two primary triplets is a single variable $i$. Based on Equation 4 and the definition of entropy, after message calibration results in Equation 7:
 \begin{align}
\begin{array}{ll}
H({{\tilde b}_i}) \propto {{\tilde b}_i} &= \sum\limits_{j,p} {{{\tilde f}_{i,j,p}}} \prod {{m_{i,j}}} \prod {{m_{i,p}}} \\
&= \sum\limits_{k,q} {{{\tilde f}_{i,k,q}}} \prod {{m_{i,k}}} \prod {{m_{i,q}}} 
\end{array}\label{equa7}
\end{align} 
where ${\tilde f_{i,j,p}},{\tilde f_{i,k,q}}$ are triplet factors associated with the two primary triplets that contain $X_i$. All pair-wise messages are incoming messages to the two primary triplets, which are sent from the child regions of the two primary triplets. \\Messages ${m_{i,p}}$ and ${m_{i,q}}$ do not result from the introduction of $\{ {X_i},{X_j},{X_k}\} $ so they are `fixed' here. The messages that vary because of the introduction of $\{ {X_i},{X_j},{X_k}\} $ are ${m_{i,j}}$ and ${m_{i,k}}$. Based on Equation 5 and 6, ${m_{i,j}}$ and ${m_{i,k}}$ are all messages originated from \underline{factor regions} to regions for node pairs $\{ i,j\} $ and $\{ i,k\} $, in the form of pair-wise and singleton messages. Among these factor regions only those contain $i$ and determine local pair-wise correlation of $\{ i,j\} $ or $\{ i,k\} $, determines $H({\tilde b_i})$ given other conditions are fixed.\\
We therefore only need to find the factors that contain $i$ and also contain variables that $j$ or $k$ depends on through $i$ (so it determines local pair-wise correlations of $\{ i,j\} $ or $\{ i,k\} $). We denote the set of these factors as ${{\bf{\Phi }}_i}$.\\
If ${{\bf{\Phi }}_i}$ is composed by the two primary triplet factors exclusively, we can reject the introduced interaction triplet, since the local pair-wise correlation over $\{ i,j\} $ or $\{ i,k\} $ are encoded already in the two primary triplets. \\
To identify ${{\bf{\Phi }}_i}$ we use Figure 6 below.
\begin{figure}[h]
		
	\includegraphics[scale=.45]{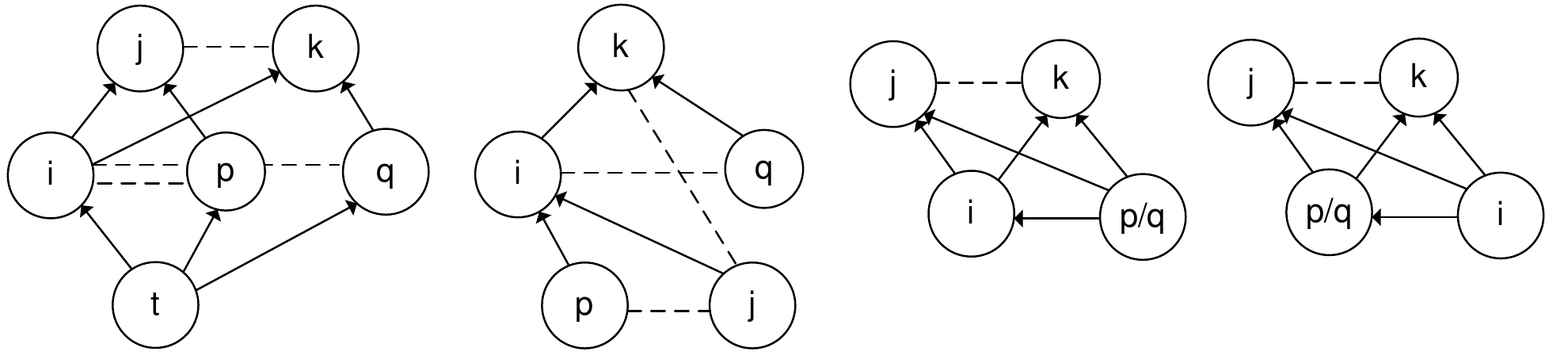}

	\qquad \quad  (a) \qquad\qquad\quad \quad \ (b) \  \qquad\qquad\quad (c)\qquad\quad \qquad  (d)
	\caption{(a) $\{ i,j,k\} $ not redundant; (b) $\{ i,j,k\} $ type 1 redundant; (c) $\{ i,j,k\} $ not redundant; (d) $\{ i,j,k\} $ type 2 redundant }
\end{figure}
\\Figure 6 illustrates partial structures of a BFG $G'$, with the node pair that is either a moral edge or that is not directly connected as an edge in $M[G']$ is marked as dashed line.\\
In Figure 6 (a), if both $j$ and $k$ are $i$'s children, ${{\bf{\Phi }}_i}$ is composed by the two primary triplet factors ${\tilde f_{i,j,p}}$, ${\tilde f_{i,k,q}}$, and other factors that contain $i$ and $pa\{i\}$, such as factor ${\tilde f_{t,i}}$ (there exists paths $t \to i \to j$ and $t \to i \to k$). So in Figure 6 (a) ${{\bf{\Phi }}_i}$ are not determined exclusively by the two primary triplet factors. Therefore regions for node pairs $\{ i,j\} $ and $\{ i,k\} $ are necessary in the region graph to capture other factor's information beside the two primary triplets; the interaction triplet cannot be rejected. \\
If $k$ is $i$'s child, $j$ must be $i$'s parent ($j$, $k$ cannot be $i$'s parent at the same time), as shown in Figure 6 (b). Here all factors that determine ${{\bf{\Phi }}_i}$ are factors in the two primary triplets. The interaction triplet is one that contains a node pair $\{ j,k\} $ that is not an edge in $M[G']$ and can be rejected (type 1).\\
Next, we consider Type 2 redundancy, which only occurs when $p=q$: \\ 
Given $p=q$, the only circumstance is when $i = 1\;or\;2$, $\{ i,j\}  \cap \{ i,k\}  \subset \{ i,j,p\}  \cap \{ i,k,q\}  = \{ {X_1},{X_2}\} $.\\
If ${X_i} = {X_2}$ (Figure 6 (c)) $X_2$ must have parent $X_1$ and apart from factors ${\tilde f_{i,j,p}}$, ${\tilde f_{i,k,q}}$, there exists a factor ${\tilde f_{{X_1}{X_2}}}$ that also determines ${{\bf{\Phi }}_i}$ and so $\{ i,j,k\} $ cannot be rejected.\\
If ${X_i} = {X_1}$ (Figure 6 (d)), $X_1$ is root node with a singleton factor ${\tilde f_{{X_1}}}$, which is always associated to $X_1$. Thus ${{\bf{\Phi }}_i}$ is determined by ${\tilde f_{{X_1}}}$ and the two primary triplet factors, which will not change by introducing $\{ i,j,k\} $. The corresponding redundant interaction triplet is then one that contains a root node and a moral edge (type 2).$\square$ \\\\
We use the graph in Figure 5 (c) to demonstrate how proposition 3 is applied.
 \begin{table}[h]
 	\caption{Redundant interaction triplet example} \label{table1}
 	\begin{center}	
 		\begin{tabular}{l| l| l|l}
 			{ Candidate }  &{ Node pairs} &{ Primary triplets} &{Redundancy}\\ 
 			\hline
 			&&&\\
 			$\{{X_1}{X_3}{X_4}\}$  &\quad  $\{{X_1}{X_3}\}$   &\quad $\{{X_1}{X_2}{X_3}\}$ & Type 1 \\
 			 &\quad  $\{{X_3}{X_4}\}$ &\quad $\{{X_3}{X_4}{E_1}\}$ \\ \hline
 		   &     &  &  \\
 			$\{{X_2}{X_3}{E_1}\}$ &\quad  $\{{X_2}{X_3}\}$ &\quad $\{{X_1}{X_2}{X_3}\}$  & N/A \\ 
 		  &\quad  $\{{X_2}{E_1}\}$     & \quad $\{{X_1}{X_2}{E_1}\}$ & \\ \hline
 		 & & & \\
 		$\{{X_1}{X_3}{E_2}\}$ &\quad  $\{{X_1}{E_2}\}$  &\quad $\{{X_1}{X_2}{E_2}\}$ & Type 2 \\
 		&\quad  $\{{X_1}{X_3}\}$ &\quad $\{{X_1}{X_2}{X_3}\}$&
 		\end{tabular}
 	\end{center}
 \end{table}
 \\Table 1 lists three candidate interaction triplets at the first level regions for a region graph built for Figure 5 (c). The related interaction regions (node pairs) and primary triplets are also listed for clarification. \\
 For example, in Table 1, for interaction triplet $\{ {X_1}{X_3}{X_4}\} $, the shared subset of both node pairs and primary triplets is the singleton $\{{X_3}\}$, and ${X_3}$ does not have any parent from other factors except these two primary triplets. So this interaction triplet is an instance identical to Figure 6 (b).\\
 Likewise, in table 1 interaction triplet $\{ {X_2}{X_3}{E_1}\} $ has node pairs $\{ {X_2},{X_3}\} $ and $\{ {X_2},{E_1}\} $ with shared subset $\{{X_2}\}$, and the primary triplets including these node pairs share a subset $\{ {X_1},{X_2}\} $. This interaction triplet cannot be rejected as it is an instance of Figure 6 (c).\\
 The interaction triplet $\{ {X_1}{X_3}{E_2}\} $ can also be rejected as it is an instance of Figure 6 (d).
 \subsection{Verification of Theorem 3 for TRC region graph}
Now all outer regions are determined by primary triplets plus interaction triplets the corresponding region graph can be generated by the CVM algorithm. The resulting region graph for our BFG models contain three levels, with all first level region counts equal to one (as all factors are included in first level). The resulting region graph is our TRC region graph which now show will satisfy both necessary conditions of Thereon 3. \\ 
In general the TRC region graph properties are summarized in Table 2 (a Proof of these results is given in Appendix A).
   \begin{table}[h]
   	\caption{Properties for $\kappa_n$ ($n>3$) dimensional $\mathcal{G}(G')$} \label{table2}
   	\begin{center}
    \small 
   		\begin{tabular}{l| l| l |l|l}
   			{\bf Levels} &{$v(r)$}  &{\it \ length} &{$max(c_r)$} &{$min(c_r)$}\\ \hline
   			&&&&\\
   			\nth{1} level  &\  3   &${(n - 2)^2}$ &\quad 1 &\quad 1\\
   			\nth{2} level  & \ 2    &${(n - 2)^2}$ &\quad -1 &\quad $3-n$ \\
   			\nth{3} level & \   1    &  $(n-3)$ & \quad $n-3$ & \quad 1
   		\end{tabular}
   	\end{center}
   \end{table}
 \\Table 2 illustrates the region size $v(r)$, the number of regions contained in each level, and $max$ and $min$ of counting numbers in each level's regions.\\\\
 Proof that the TRC region graph satisfies the perfect correlation property \cite{YedidiaFW05}:\\
 \textit{Proof.} Based on Table 2 (and Appendix A), the sum of all first level region counts is ${(n - 2)^2} \times 1 = {n^2} - 4$. The second and third level regions' counts are cancelled by each other, which will leave one region with counting $3-n$ (there are two regions at the second level with counting $3-n$ and one is cancelled) and ${(n - 2)^2} - (n - 3) - 1$ regions with counting $-1$. So, sum them all to obtain ${n^2} - 4 + 3 - n + ({(n - 2)^2} - (n - 3) - 1) \times  - 1 = 1$.$\square$\\\\
 Proof that the TRC region graph satisfies maxent-entropy normal property \cite{YedidiaFW05}: \\
 \textit{Proof}. The Bethe approximation is maxent-normal \cite{YedidiaFW05}, and so the entropy of the region graph, ${H_{\cal G}}$, can be written as ${H_{\cal G}} = \sum\limits_{i = 1}^N {H({b_i})}  - \sum\limits_{a = 1}^M {I({b_a})} $ where $N$ is the number of variables in the region graph, $X_i$, and $M$ is the number of factors, $a$, (${{\bf{x}}_a}$ are the variables defined by the factor $a$). $H({b_i}) \equiv  - \sum\limits_{{x_i}} {{b_i}({x_i})\ln {b_i}({x_i})} $ s the sum of entropies from all variables $X_i$ in the region graph, and $I({b_a}) \equiv \sum\limits_{{{\bf{x}}_a}} {{b_a}({{\bf{x}}_a})\ln {b_a}({{\bf{x}}_a})}  - \sum\limits_{i \in N(a)} {H({b_i})} $ is the mutual information which is the entropy for a region containing factor $a$, minus the entropies of all variables contained in factor $a$. ${H_{\cal G}}$ is maximal, equalling $\sum\limits_{i = 1}^N {H({b_i})} $, when all beliefs, ${b_i}({x_i})$ and ${b_a}({{\bf{x}}_a})$, are uniform, and under these circumstances the mutual information, $I({b_a})$, equals zero. In our region graph we can always construct ${H_{\cal G}}$ in the form of ${H_{\cal G}} = \sum\limits_{i = 1}^N {H({b_i})}  - \sum\limits_{a = 1}^M {I({b_a})} $ because the mutual information for each triplet can be constructed by its connected second level regions and the single variables the triplet contains, resulting in minimal $I$ terms and maximal entropy ${H_{\cal G}}$ when all beliefs are uniform. The rest of the proof is omitted for brevity because the verification can be done directly on the TRC region graph.$\square$ \\\\
An example of TRC region graph for Figure 5 (c) is shown in Figure 7.
\begin{figure}[h]
	
	\includegraphics[scale=.5]{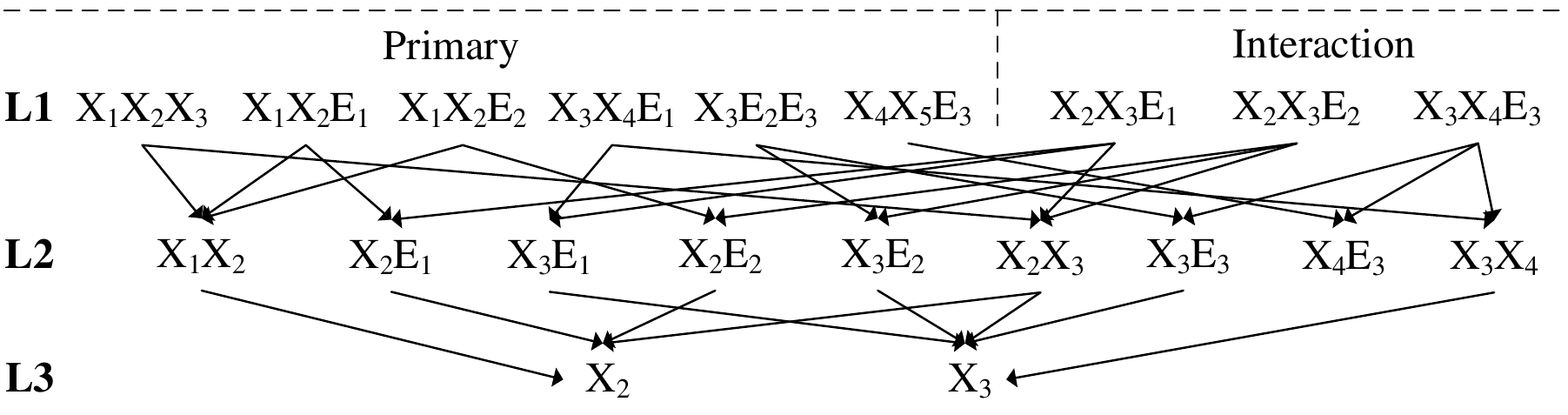}
	
	\caption{TRC region graph for Figure 5 (c)}
\end{figure}    
 
 \subsection{Region Graph Binary Factorization Algorithm}
 To use message updating equations on the region graph we need to solve the numerical instability problem discussed in section 3.2. Recall that ${c_r} = 1 - \sum\limits_{r' \in Ancestor(r)} {{c_{r'}}} $, and so a large absolute value of counting number also implies global multiplicity of connected regions. In Table 2, the connections between first and second level regions grow because the  $min(c_r)$ is linearly decreasing (conversely $max(c_r)$ is linearly increasing), which means the number of multiple connections grow and we are guaranteed to encounter a numerical instability problem from multiple cycles in the region graph. This is also evident in the example we have used in Figure 7. The CCCP algorithm suffers from the same problem. To reduce the absolute value of the counting number and decompose the multiple connections within a region graph we use the following RGBF algorithm.\\ \\ 
 \textbf{Definition 8.} \textit{A Region Graph Binary Factorization (RGBF) algorithm is one that ensures that each region in a region graph, originally with more than two parents, has exactly two parents without changing the validity of a region graph.  }\\ \\
 The particular RGBF that we propose is described in Algorithm 2. This RGBF algorithm will be used to generate an equivalent region graph ${\cal G}'$ from the original region graph ${\cal G}$, with the properties described in the following proposition. \\\\
 \textbf{Proposition 4.} By applying the RGBF of algorithm 2 we transform a $k$-level CVM region graph ${\cal G}$ with all factors included in the \nth{1} level, into an \underline{equivalent} $k$-level region graph ${\cal G}'$, such that each region $r$ in ${\cal G}'$ ($r \in R, r \notin {R_{\nth{1}level}}$) is connected to two parents. The counting numbers for all regions are 1, -1 and 0. \\\\
 \textit{Proof.} Algorithm 2 will produce ${p_r} - 1$ (${p_r}$ the number of parents) copies of region $r$ in ${\cal G}$ to ${\cal G}'$ when ${c_r}$ is not 1, -1 or 0 in ${\cal G}$. Each $r$ region in ${\cal G}'$ will share one parent with its neighbouring $r$ copy. Equivalence between ${\cal G}$ and ${\cal G}'$ can be proved by satisfying the consistency and unity conditions \footnote{Unity is defined as where the sum of all regions counting numbers associated with each variable should be one.} for a region graph. 
 \begin{enumerate}
 \item Consistency: as the first level is not changed, consistency of all $r$ $(r \notin {R_{1st\;level}})$ and its copies with their parents in ${\cal G}'$ must be maintained. This is satisfied as each $r$ is connected with its neighbouring copy by sharing one parent, so all parents and all regions $r$ are connected and hence consistent. 
\item Unity: global unity for each variable must be the same in ${\cal G}$ and ${\cal G}'$. As ${\cal G}'$ does not contain any new regions compared to ${\cal G}$ but only copies of regions, $r$, from ${\cal G}$, the counting number for each variable will only be influenced by the region $r$ and its copies. So the unity condition can be satisfied by integer accumulation of $r$ and its copies' counting numbers in ${\cal G}'$ to ${c_r}$ in ${\cal G}$, $\sum\nolimits_{i = 1}^{{p_r} - 1} {{c_{{r_i}}}}  = {c_r}({r_i} \in {\cal G}',r \in {\cal G})$, which will not change the unity condition for each variable. In this way the cumulative counting number is not unique but can be specified by using 1, -1 and 0 as these work for any integer. $\square$ 
 \end{enumerate}
  \begin{algorithm}[!htbp]
  	\SetCommentSty{small}
  	\caption{RGBF algorithm}
  	\KwIn{$k$-level CVM region graph ${\cal G}$ with regions $R$ }
  	\KwOut{$k$-level region graph ${\cal G}'$}
  	\textbf{Initialize:} \ ${\cal G}' \leftarrow \emptyset $\;
  	\qquad\quad\qquad ${\cal G}' \leftarrow {\cal G}' \cup {\cal G}_{\nth{1}level}$\;

  	\For{$i=2:k$ }
  	{
  		\For {each region $r \in {R _{i^{th}level}}$ }{
  			
  			\If{$p_r >2$}{
  				
  			\For{$z=1:p_r-1$}{
  				$r'_z \leftarrow$ repeat $r$ \\
  				connect $r'_z$ to two $parent(r)$\\
  				$c_{r'_z} \leftarrow$ cumulative integer total $c_r$
  				}
  				$  \mathcal{G}_{i^{th}level}$ replace $r$ by $r_1',...,r_{p_r-1}'$ 
  				}

  		}
  		${\cal G}' \leftarrow {\cal G}' \cup  \mathcal{G}_{i^{th}level}$
  	}
  	return ${\cal G}'$\;
  \end{algorithm}
The benefit of applying the RGBF algorithm on a region graph is that large counting numbers no longer occur and multiple connections are decomposed into local connections. Therefore, the number of cycles in the region graph is reduced to a minimal number and each third level region will be associated with at most one cycle. An example is shown in Figure 8.
\begin{figure}[h]
	\vspace{.001in}
	\centering
	\includegraphics[scale=.5]{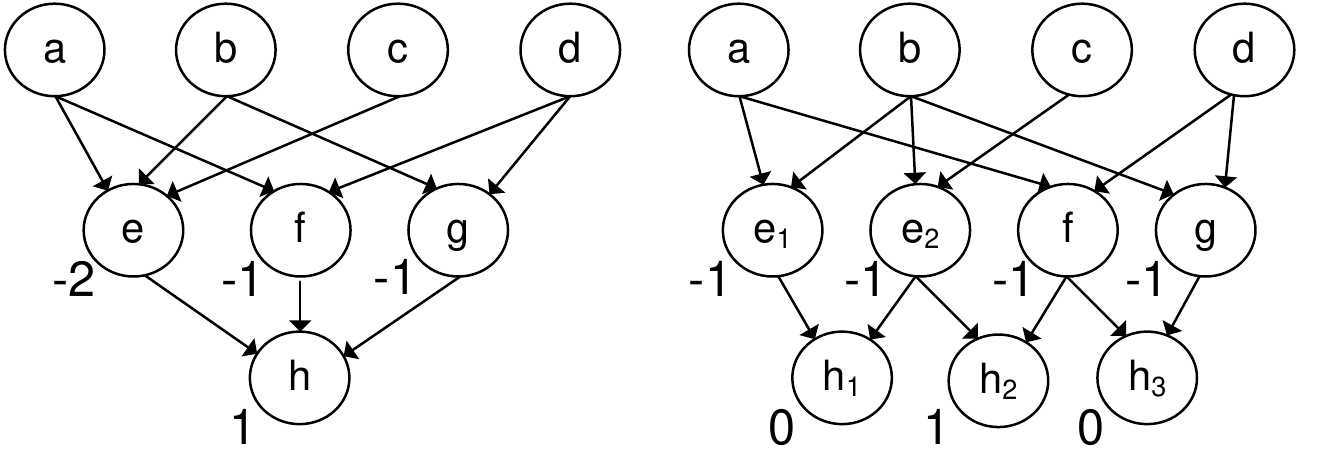}
	\\
	(a) \quad  \qquad  \qquad\qquad\qquad    (b)
	\caption{(a) region graph $ \mathcal{G} $ (all \nth{1} level regions' counting numbers 1); (b) region graph $ \mathcal{G'} $ by RGBF process of (a)}
\end{figure} 
\\In Figure 8, regions $e$ and $h$ are copied twice and three times respectively. The counting numbers for each \nth{2} level region becomes -1, and for each \nth{3} level region becomes 0 or 1. It does not matter if 1 is placed on $h_1$ or $h_2$ since it does not change the consistency and unity conditions, but it will influence the convergence speed. The RGBF process does not change the consistency and unity conditions indicated by $\mathcal{G}$. If we used the GBP algorithm there would be a limited number of messages multiplied into equations 4 and 6 at each updating iteration, and there is no large counting number for calculating $b_r$. \\ 
If we use RGBF algorithm for a multiply connected CVM region graph, the CCCP updating will also be robust.\\ \\
\textbf{CCCP update rule:} \cite{Yuille02cccpalgorithms}
\begin{align}
{h_r}({x_r}) = {e^{ - \frac{{{c_r}}}{{{c_{\max }}}}\{ {E_r}({x_r}) + 1\} }}{\{ {b_r}({x_r})\} ^{\frac{{{c_{\max }} - {c_r}}}{{{c_{\max }}}}}}\label{equa8}
\end{align}
\begin{align}
{g_r}({x_r}) = {e^{ - {\gamma _r} - \sum\limits_{s \in child(r)} {{\lambda _{r \to s}}({x_s})}  + \sum\limits_{v \in parent(r)} {{\lambda _{v \to r}}({x_r})} }}\label{equa9}
\end{align}
\begin{align}
{b_r}({x_r}) = {h_r}({x_r}){g_r}({x_r})\label{equa10}
\end{align}
\begin{align}
{e^{2{\lambda _{r \to u}}({x_u};\tau  + 1)}} = {e^{2{\lambda _{r \to u}}({x_u};\tau )}}\frac{{\sum\nolimits_{x \in r\backslash u} {{b_r}} }}{{{b_u}}}
\label{equa11}
\end{align}
where $\lambda$ and $\gamma$ are parent-child region consistency and normalization Lagrangian multipliers.
 $c_{max}$ is the max value of all regions' counting numbers in a region graph. 
 $h_r$ and $g_r$ are pre-calculated parameters for computing belief term $b_r$. \\\\
  In the CCCP algorithm, updating each $\lambda_{r \to u}$ is a recursive process that involves calculating the beliefs over all $u$'s parents and children, and its children's parents. In Figure 8 (a) updating $\lambda_{a \to e}$ involves the belief calculations of seven regions at a time: $a$, $b$, $c$, $e$, $f$, $g$ and $h$. This number grows with the multiple connections for $h$ and the number of cycles associated with $h$ also grows (there are three cycles associated with $h$ in Figure 8 (a)). But after applying RGBF to (a), as shown in (b), to update $\lambda_{a \to e_1}$ there are now five regions ($a$, $b$, $e_1$, $e_2$, $h_1$) and this number does not increase with the number of connections because there are no multiple connections and only one cycle, maximum, for each level three region.
  \subsection{TRC Algorithm and Its Complexity}
  \begin{algorithm}[!htbp]
  	\SetCommentSty{small}
  	\caption{TRC algorithm}
  	\KwIn{$\kappa_n$ dimensional BFG $G'$,\ $\varepsilon  = 1.0e - 5$}
  	\KwOut{$G'$ with marginal distributions}
  	\textbf{Initialize:} $M[G']$ $\leftarrow$ parametrizing BN $G'$ to MN\;
  	\qquad\qquad\quad \ $\mathcal{F} \leftarrow \phi_j (\phi_j \in p')$; $\mathcal{U} \leftarrow \emptyset$\;
  	\For{moral edge $\mathcal{M}_t = (X_i,E_t), (\mathcal{M}_t \in M[G'])$}
  	{
  		${\cal U} \leftarrow {\cal U} \cup \{ {{\cal M}_t} \cup {X_i}\} $ $({X_i} \in \partial {{\cal M}_t})$\;
  	}
  	Drop redundancy in ${\cal U}$\;
  	$\mathcal{G} \leftarrow CVM(\{ \mathcal{F} \cup \mathcal{U} \}) $\;
   
  		$\mathcal{G'} \leftarrow Algorithm2(\mathcal{G})$\;
  	 
  	\textbf{parallel} \For{$r$ ($r \in {R_{\nth{1}level}}, {R_{\nth{1}level}} \in \mathcal{G'}$)}{
  		\If{$|b_r^{old} - {b_r}| > \varepsilon $}{
  			$b_r \leftarrow equation \ref{equa10} $\\
  			\If{$child(r) \neq \emptyset$}{
  				$b_{u \in child(r)} \leftarrow recursion(equation \ref{equa10})$\\
  			}
  			$\lambda_{r \to u} \leftarrow equation \ \ref{equa11}$}}
  	return $G'$\;
  \end{algorithm}
 For convenience we can use the CCCP update rule in parallel because to update each $\lambda$ there are only limited regions involved for computation, which is a result of the RGBF algorithm. For example, we can update $\lambda_{r \to u}$ simultaneously when updating $\lambda_{i \to j}$ provided that $i \neq j \neq r$ and $i \neq j \neq u$. Convergence is guaranteed and is determined by the discrepancy between old and current beliefs.  
  The clustering complexity of TRC is the sum of all levels' regions; this is polynomial and proportional to $\sum\nolimits_{3\ levels} {v(r) \cdot length} $ (as shown in Table 2) in contrast to exponential clustering complexity for exact methods. Computational complexity is proportional to the number of \nth{1} to \nth{2} level region edges, which is the sum of all \nth{2} level's degree of freedom, $\sum\limits_{j = 1}^{{{(n - 2)}^2}} {(|{c_{{r_j}}}| + 1)} $ and is polynomial. A proof of these results is given in Appendix A. Efficiency can be further improved by using parallel processing. 
  \section{Experiments}
  This section presents experiments conducted to determine how well TRC performs for general and high tree-width BFG models compared with competing methods. Typically, experiments carried out in the literature have focused on spin glass models  \cite{YedidiaFW05,Kolmogorov:2006:,Sun:2003:SMU}. but we did not use these because they are undirected models and we are interested in directed models. Also, with spin class models the counting number is relatively small and the number of multiple connections is low compared to BFGs. This means spin glass models are easier test cases than BFGs with respect to the numerical instability problem and so we concentrate on experiments involving more challenging BFGs.\\
  Section 5.1 presents the results of testing two simple sparse BN models and compares TRC with the exact and MCMC solutions. Note that, in contrast with MCMC, TRC is not problem tailored and does not need any parameter adjustment. \\
  Section 5.2 presents the results of testing a number of high tree-width BFG models with different numbers of discrete states to investigate efficiency, robustness and accuracy and compare it with a JT solution.\\
  Obviously the RGBF algorithm is an independent step that can be used or not. So in Section 5.3 we compare the results by switching RGBF on/off for both the GBP and CCCP algorithms and find that the RGBF process improves the stability of both algorithms.\\
  Generally, the efficiency of TRC depends on which underlying message update rule is used, either GBP or CCCP or other similar rules, whereas CCCP is slower than GBP [20]. In our tests models of less than 12 dimensions compute within one minute. Computation time for 100 dimensions is up to three hours.\\
  The environment for testing was Java JDK 1.8, Intel E2660 @ 2.2 GHz. 
  
 \subsection{Sparse BN Model Test}
  \begin{table}[h]
  	\caption{Mean value of Asia model variables' marginal results, given variable $a=2$ and $d=2$ }  
   
  	\begin{center}
  		\begin{tabular}{l| l| l |l|l|l|l|l}
  			Thre.&{Method} &{$s$}  & $t$ &{$l$} &{$b$} &{$e$} &$x$ \\ \hline
  			&&&&&&&\\
  			&Exact  & 1.626   & 1.088 & 1.100 & 1.811 & 1.182 &1.220\\
  			1.0e-5&TRC  & 1.626   & 1.088 &1.100 & 1.811& 1.183 & 1.220\\
  			1.0e4&MCMC & 1.630  &  1.084  & 1.107 & 1.816 & 1.185 & 1.222\\
  			1.0e5&MCMC & 1.626 & 1.086 & 1.100 & 1.815 & 1.181 & 1.218
  		\end{tabular}
  	\end{center}
  \end{table}
  We present the well known Asia model for illustration of a sparse BN graph model by using the TRC algorithm. We change the state names to ``1'' and ``2'' to accommodate the result shown in WingBugs \cite{winb}. We have restricted our analysis throughout this paper to complete BNs (for reasons explained in Section 2), but as explained in Section 2 we can always convert a sparse BN into an associated complete BN before applying the BFG process. Hence for sparse BNs to use TRC we use the BFG from its associated complete BN. (see Appendix B for the details).\\
   The TRC result of the Asia model (Table 3) is obtained using a convergence threshold of 1.0e-5. The MCMC threshold, sample size, is increased from 1.0e4 to 1.0e5. Note that in this test TRC is more accurate than MCMC, but more importantly is guaranteed to converge.  
  \begin{figure}[h]
  	\vspace{.001in}
  	\centering
  	\includegraphics[scale=.6]{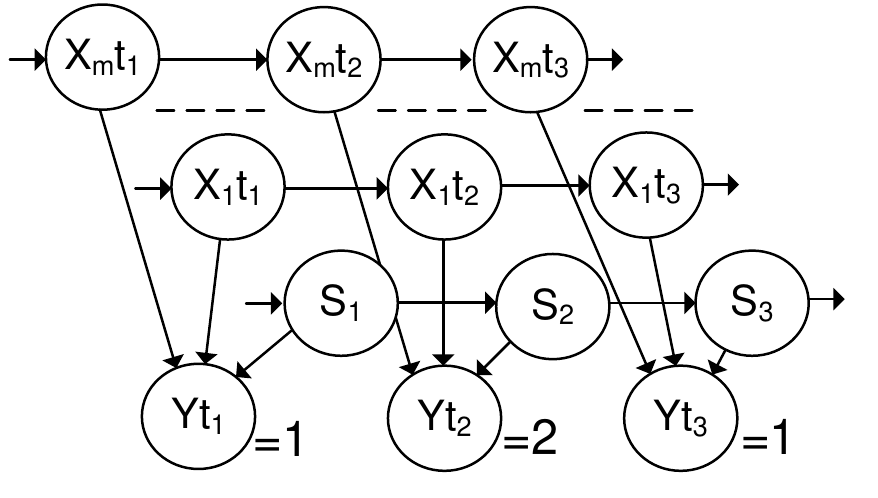}
  	\caption{a Dynamic Bayesian Network (DBN) containing three time slices}
  \end{figure}
 \\Using the same approach we also tested the sparse BN model shown in Figure 9. This is a Dynamic Bayesian network (DBN), in the form of the switching state space model \cite{Ghahramani:2001}. DBN is a popular class of model used in time series analysis, including robotics, protein sequencing and many other domains. \\
 We assume all variables are discretized and are binary variables. Although the model is a sparse BN the tree-width grows with the value of $m$, where ${X_m}$ is a hidden vector node. So the exact solution may be intractable when $m$ is large. We set observed values for output node $Y$ in this test. In TRC the clustering complexity remains polynomial regardless of the value of $m$.  
   \begin{table}[h]
   	\caption{Mean value of Fig 9 marginal results (with all binary variables and states ``1'' and ``2'', NPT setting is in Appendix B)}  
   	\begin{center}
   		\begin{tabular}{l| l| l |l|l|l|l|l}
   			 Method &{$s_1$}  & $s_2$ &{$s_3$} &{${x_1}{t_2}$} &{${x_1}{t_3}$} &${x_m}{t_1}$ & ${x_m}{t_3}$ \\ \hline
   			&&&&&&&\\
   				Exact  & 1.408   & 1.782  & 1.113 & 1.521  & 1.797  & 1.756 & 1.396\\
   			 TRC  & 1.408   & 1.781  & 1.114  & 1.521 & 1.796  & 1.756 & 1.396\\
   			 MCMC & 1.407  &  1.776  & 1.112 & 1.517 & 1.796  & 1.762 & 1.401\\
   			 MCMC & 1.407  &  1.782   & 1.114  & 1.521  & 1.795 & 1.758 & 1.396
   		\end{tabular}
   	\end{center}
   \end{table}  
 \\Comparison of the exact, TRC and MCMC results as shown in Table 4 (variables ${x_1}{t_1}$ and ${x_m}{t_2}$ are not listed as all three methods have the same results). MCMC sample sizes are 1.0e4 (the upper row) and 1.0e5 (the lower row). The TRC convergence threshold is 1.0e-5. TRC approximates all variables very well, achieving a maximum relative error of 0.001 to exact values. Notice that the MCMC result is improved by increasing the number of samples (from 1.0e4 to 1.0e5), and it achieves a maximum relative error of 0.002 with exact values under the threshold of 1.0e5.    
\subsection{BFG (from complete graph) Models Test}  
\begin{figure}[h]
	\vspace{.001in}
	\centering
	\includegraphics[scale=.75]{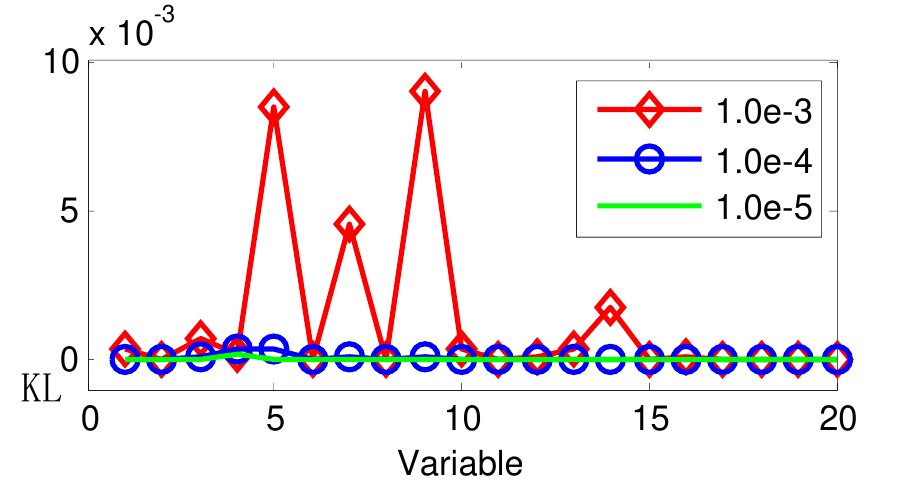}
	\caption{$\kappa_{20}$ BFG (binary variables with random NPTs) test of TRC using different convergence thresholds}
\end{figure}  
We first present a ${\kappa _{20}}$ BFG model test by using different convergence thresholds. In Figure
10 we change the convergence threshold $\varepsilon $ from 1.0e-3 to 1.0e-5. For each threshold the graph shows the accuracy (measured by the KL distance) compared to the exact solution. As the convergence threshold increases from 1.0e-3 to 1.0e-5 the accuracy increases. When $\varepsilon  = 1.0e - 5$ the maximum KL distance is below 1.6e-4, indicating effective improvement of accuracy by increasing the convergence threshold. 
\begin{table}[h]
	\caption{Space complexity and KL distance of TRC for high tree-width BFG models (binary variables with random NPTs) test} \label{table5}

	\begin{center}
		\begin{tabular}{l|l|l|l|l}
		     &\multicolumn{4}{c}{\bf{Tree-Width}} \\
			 &{19 ($\kappa_{20}$)}  &{39 ($\kappa_{40}$)} &{79 ($\kappa_{80}$)}  &{99 ($\kappa_{100}$)}\\ \hline
			
		Space complexity	&&&&\\
		\quad	JT $O(2^n)$ & 8 Mb  &  8e3 Gb& 9e15 Gb  & 9e21 Gb\\
		\quad	TRC $O(n^2)$ & .06 Mb &  .11 Mb & .47 Mb & .73 Mb\\
			\qquad (iterations) &\quad (782)   &\quad (1567) &\quad (3561)  & \quad (3963)\\
			\hline
		
			 	KL for TRC	&&&&\\
		\quad	$max(KL)$ & 1.53e-4  & 1.9e-5 & 3.8e-5 & 2.8e-5\\
		\quad	$min(KL)$  & 3.65e-12    & 2.5e-13 & 2.1e-9  & 4.5e-8 \\
		\quad	$average(KL)$  & 1.46e-5     & 5.2e-6 & 7.5e-6  & 2.9e-6 
		\end{tabular}
	\end{center}
\end{table}
\\Table 5 is a summary of the test results for $\kappa_{20}$, $\kappa_{40}$, $\kappa_{80}$ and $\kappa_{100}$ respectively (with $\varepsilon  = 1.0e - 5$), tree-width parameter for each model is measured by JT solution. $O{(2^n})$ and $O({n^2})$ are space complexities.
These results show that the clustering complexity is reduced from exponential to polynomial (from gigabytes to less than 1 megabyte). As the dimensions increase the accuracy does not notably decrease; all KL statistics show a robust and accurate performance and we can increase the convergence threshold to obtain higher accuracy. Because exact computation for all variables is not possible with finite memory, we could not compute exact values beyond 25 dimensions. Therefore, we compare the accuracy of the first 20 dimensions produced under TRC, for all sizes of models, with these exact values and infer that if the TRC results are accurate for these first 20 then the other variables, that could not be directly compared, must be accurate too (because the model has converged).\\ 
  \begin{table}[h]
  	\caption{$\kappa_{20}$ BFG model tests with random NPT and different number of discrete states, with convergence threshold set by 1.0e-5 except the last column} \label{table6}
    	\begin{center}
  		\begin{tabular}{l| l| l |l|l|l}
  			 &\multicolumn{5}{c}{\bf{Discrete state}} \\
  		 &{$m=3$}  & $m=4$ &{$m=5$} &{$m=6$} &{$m=6$}\\ 	&&&&& (1.0e-6)\\   \hline
  			&&&&&\\
  			Iterations & 860   & 642 & 705 & 588 & 1048 \\
  			\hline
  				&&&&&\\
  				KL for TRC &&&&&\\
  		\quad	$max(KL)$ & 1.15e-5  &  1.42e-5  & 1.19e-5 & 3.14e-5 & 5.87e-6 \\
  		\quad	$min(KL)$ & 4.71e-8  &  1.35e-8  & 3.18e-8 & 1.64e-7 & 2.76e-9 \\
  		\quad	$average(KL)$ & 2.99e-6  &  3.9e-6  & 3.31e-6 & 7.25e-6 & 1.12e-6 
  		\end{tabular}
  	\end{center}
  \end{table}
 \\We compared the TRC result with the exact value for 1 to 10 dimensions for each test model and discovered that as the number of discrete states $m$, for each variable, increases, the number of iterations required to converge increases in general but also depends on BN parameterizations. This is shown in Table 6, where when convergence threshold is 1.0e-5, the KL statistics degrade slightly with the increase in the number of states, but when the convergence threshold is set to 1.0e-6, the KL is reduced. So, as the number of discrete states increases we can set a higher converge threshold to guarantee accuracy. 
\subsection{RGBF Test} 
  \begin{figure}[h]
  	\vspace{.001in}
  	\centering
  	\includegraphics[scale=.5]{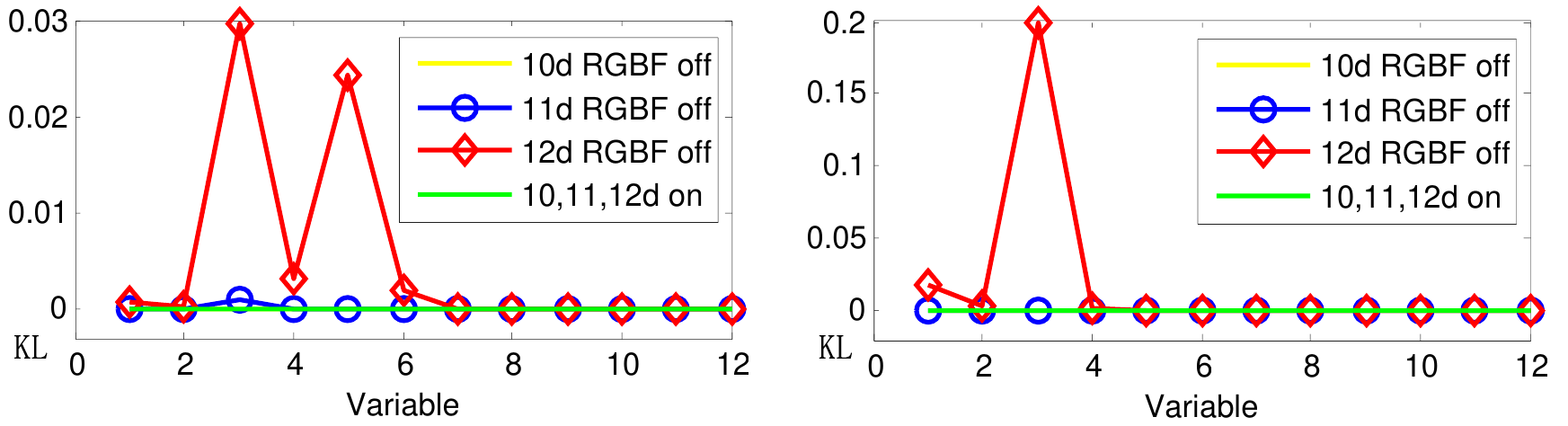}
  	\\
  	(a) \qquad \qquad  \qquad\qquad\qquad \qquad   (b)
  	\caption{(a) GBP with/without RGBF on $\varepsilon  = 1.0e-5$; (b) CCCP with/without RGBF on $\varepsilon  = 1.0e-5$}
  \end{figure} 
  To investigate the effectiveness of the RGBF algorithm, Figure 11 shows the results of using the GBP and CCCP inference algorithms (showing the problem occurring after $n=10$) with and without RGBF. Without RGBF both algorithms demonstrate significant inaccuracy (for GBP these are evident in the lowest seven dimensions, and for CCCP these are evident for the lowest four dimensions). In contrast, both algorithms are accurate in all cases with RGBF. These tests also show that the low dimensional variables of a BFG model are more likely to experience numerical problems than high dimensional variables, since low dimensional variables are connected to more children than than high dimensional variables.
  \section{Conclusion and Future Work}
  We have presented a general purpose approximate Bayesian Network inference algorithm-Triplet Region Construction (TRC) that overcomes the computational complexity barrier of exact algorithms (such as Junction Tree). Specifically, whereas exact algorithms are worst case exponential for BNs with large numbers of densely connected variables (since clustering complexity can grow exponentially with the number of nodes) the TRC algorithm reduces the clustering complexity from worst case exponential to polynomial for factorized models. Likewise, the computational complexity is polynomial and can be further speeded up by parallel processing.\\
  The TRC algorithm is based on a binary factorization algorithm and composed of three sub algorithms (ORI, RGBF and CCCP) that provide systematic improvements to previous methods of region based approximate belief propagation (namely relating to region choice, convergence and accuracy). 
  The ORI and RGBF algorithms are independent for high dimensional model problems and can be applied separately to many other types of models.\\ 
  Experiments carried out by using synthetic data without extreme conditional probability (probability near zero \cite{Dagum19971}), have shown that TRC is accurate and robust, and so can be used as an alternative to the JT algorithm for handling high dimensional (also high tree-width) problems. Unlike MCMC solutions, TRC is guaranteed to converge and does not require special considerations of parameter adjustment for any discrete models. \\
  Future extensions of this work will focus  on using TRC for high tree-width model parameter learning and sensitivity analysis. We will also combine TRC with discretization \cite{KozlovK97,NeilTM07} or sampling for continuous variables so that TRC can take place for all kinds of distributions
  \footnote{There are preliminary works available in \cite{Lin15e}.}.

\appendices
\section{Proofs}
\textit{Proof.} of the results in Table 2 of the paper.
\begin{enumerate}
\item Let $n$ be the number of original nodes in a BFG, $G'$, so the number of intermediate nodes in $G'$ is: $1 + 2 + ... + n - 3 = (n - 2)(n - 3)/2,\;n > 3$.
\item From the parent to child relationships in $G'$, the number of primary triplets is determined by the sum of the number of original variables and intermediate variables minus 2, as there are two factors absorbed in triplets. So we have $n - 2 + (n - 2)(n - 3)/2$ primary triplets.
\item The number of interaction triplets is the number of moral edges and it is also the number of intermediate nodes, so we have $(n - 2)(n - 3)/2$ interaction triplets. 
\item The number of first level triplets is then: ${L_1} = n - 2 + (n - 2)(n - 3)/2 + (n - 2)(n - 3)/2 = {(n - 2)^2}$.
\item The number of second level intersections is determined by the number of first level triplets and is ${(n - 2)^2}$.
\item There are $n - 3$ intersections with the form ${X_i}{X_j}$ which has counting number -1 to $3-n$, so $\min ({c_r}) = 3 - n$ at the second level. All other intersections with the form ${X_i}{E_t}$ have counting number -1.
\item The third level regions are all single variable regions and are original variables $X_i$, with the counting number 1 to $n-3$ sequentially, so $\max ({c_r}) = n - 3$. $\square$
\end{enumerate}
\textit{Proof.} of TRC complexity\\
For all BFG BNs we considered, space complexity is proportional to the sum of all the levels' regions in a TRC region graph. When RGBF is not used, the space complexity is proportional to $\sum\nolimits_{3{\kern 1pt} levels} {v(r) \cdot length} $ (as shown in Table 2). When RGBF is used, RGBF results in a linear expansion of second and third level regions so the overall space complexity remains polynomial and proportional to $\sum\nolimits_{3{\kern 1pt} levels} {v(r) \cdot length} $. \\
Efficiency complexity is proportional to the number of first to second level region edges, which is the sum of all second level's degree of freedoms $\sum\limits_{j = 1}^{{{(n - 2)}^2}} {(|{c_r}| + 1)} $ and it is polynomial. RGBF also results in a linear expansion of second level regions, and each second level region has at most four child regions by RGBF, so the second to third level region edges are linear expansions with the number of second level regions. Hence it is linearly proportional to the number of second level regions. Complexity is then determined by the number of first to second level region edges and is still proportional to $\sum\limits_{j = 1}^{{{(n - 2)}^2}} {(|{c_r}| + 1)} $. $\square$
\section{Examples}

 We can add edges to convert a sparse BN to a complete BN before applying the BF process, or alternatively, if the original sparse BN is already a subset of a BFG (i.e. each node has less than two parent nodes) we can also convert this subset of BFG $G$ directly to a BFG $G'$ by using these steps. 
 \begin{enumerate}
 	\item The number of original variables of a corresponding BFG $G'$ equals the number of all variables in  $G$. So if $G$ contains $n$ variables the corresponding $G'$ is a $\kappa_n$ BFG, in which there are $n$ original variables $\{ {X_1},...,{X_n}\} $. So there exists a unique parent-child path in $G'$ that contains all original variables of $G$, which has a parent-child ordering ${\pi _{G'}} = :\{ {X_1} \to ,...,\to {X_n}\} $.
 	\item Define a parent- child ordering ${\pi _G}$ for the $n$ original variables in $G$, in which for any $X_j$ that is successor to $X_i$ ($i,j \in n$), ${X_j} \notin pa\{ {X_i}\} $.
 	\item Ensure ${\pi _{G'}} = {\pi _G}$ in $G'$.
 	\item Define NPT for each node $X_i$ (${X_i} \in {\pi _{G'}}$) by reusing the NPT from $G$, whilst maintaining the CI information encoded in $G$. 
 \end{enumerate}
 To reuse the NPT for $X_i$ (${X_i} \in {\pi _{G'}}$) from $G$ the parent nodes $pa\{ {X_i}\} $ in $G'$ must be the same as those in $G$. This is achieved by replicating the original variables through intermediate variables in $G'$ if $pa\{ {X_i}\} $ between $G$ and $G'$ differ. \\ 
 For example, the Asia model, as shown in Figure 12 (a), is already a subset of a BFG. 
  \begin{figure}[h]
  
  	\includegraphics[scale=.42]{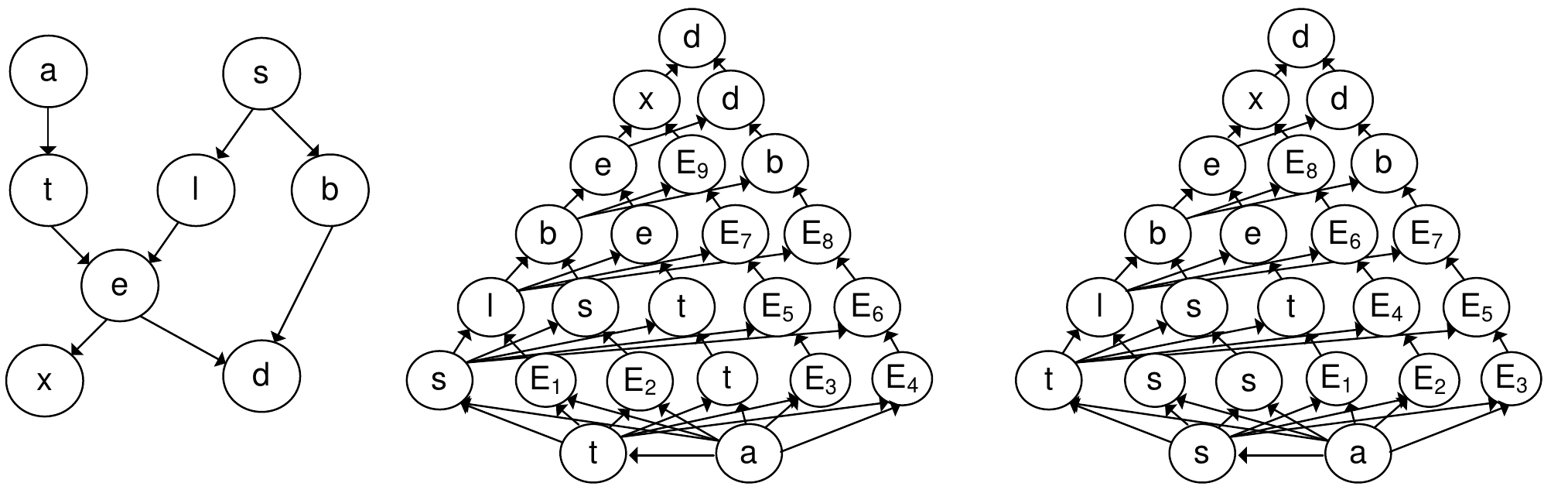}
  
  	\quad \quad \ (a) \ \qquad \qquad  \quad\quad \ \  (b)\qquad\qquad \quad \qquad\qquad (c)
  	\caption{(a) Asia model $G$; (b) correct ${\kappa _8}$ $G'$; (c) correct ${\kappa _8}$ $G'$ }
  \end{figure}
\\For example during the conversion from Figure 12 (a) to (b), the NPT, $P(e|t,l)$ in (a) is reused in (b) as $pa\{ e\} $ are the same between (a) and (b). CI information in (a) is also maintained in (b), such as where NPT $P(l|s,{E_1})$ is set as $P(l|s,{E_1}) = P(l|s)$. Here $l$ is independent with $E_1$ in (b) which is identical to (a).\\ 
Because ${\pi _{G'}}$ is not unique we can also obtain alternatives such as that shown in (c). The difference between Figure 12 (b) and (c) for ${\pi _{G'}}$ is that in (b) $a \to t \to s$ and in (c) $a \to s \to t$. Although the two alternatives differ both of them contain identical CI information as encoded in the original BN, such as in (b) $P(s|t,a)=P(s)$, and in (c) $P(t|s,a)=P(t|a)$. The marginals that result from the two alternatives are the same.\\
As there are intermediate variables that are used to replicate the original variables, when setting evidence on the original variable in $G'$ the evidence need also to be set on these replicating variables (marked with the same name as the original variables in (b) (c)).\\\\
NPT setting for Table 4 \\
p.s1= c(0.8,0.2),\\
p.s2$|$s1= c(0.1,0.9,0.2,0.8), \\
p.s3$|$s2=c(0.9,0.1,0.7,0.3), \\
p.x1t1= c(0.6,0.4),\\
p.x1t2$|$x1t1=c(0.7,0.3,0.6,0.4), \\
p.x1t3$|$x1t2= c(0.1,0.9,0.4,0.6), \\
p.xmt1= c(0.3,0.7),\\
p.xmt2$|$xmt1= c(0.2,0.8,0.3,0.7),\\
p.xmt3$|$xmt2 = c(0.4,0.6,0.7,0.3),\\
p.yt1$|$s1,x1t1,xmt1 =\\
 c(0.1,0.9,0.2,0.8,0.3,0.7,0.4,0.6,0.5,0.5,0.6,0.4,0.7,0.3,0.8,0.2),\\ 
p.yt2$|$s2,x1t2,xmt2 = \\
c(0.3,0.7,0.4,0.6,0.5,0.5,0.6,0.4,0.7,0.3,0.8,0.2,0.9,0.1,0.1,0.9), \\
p.yt3$|$s3,x1t3,xmt3 = \\
c(0.6,0.4,0.7,0.3,0.8,0.2,0.9,0.1,0.1,0.9,0.2,0.8,0.3,0.7,0.4,0.6) \\
\begin{figure}[tb]
 	\vspace{.001in}
 	\includegraphics[scale=.25]{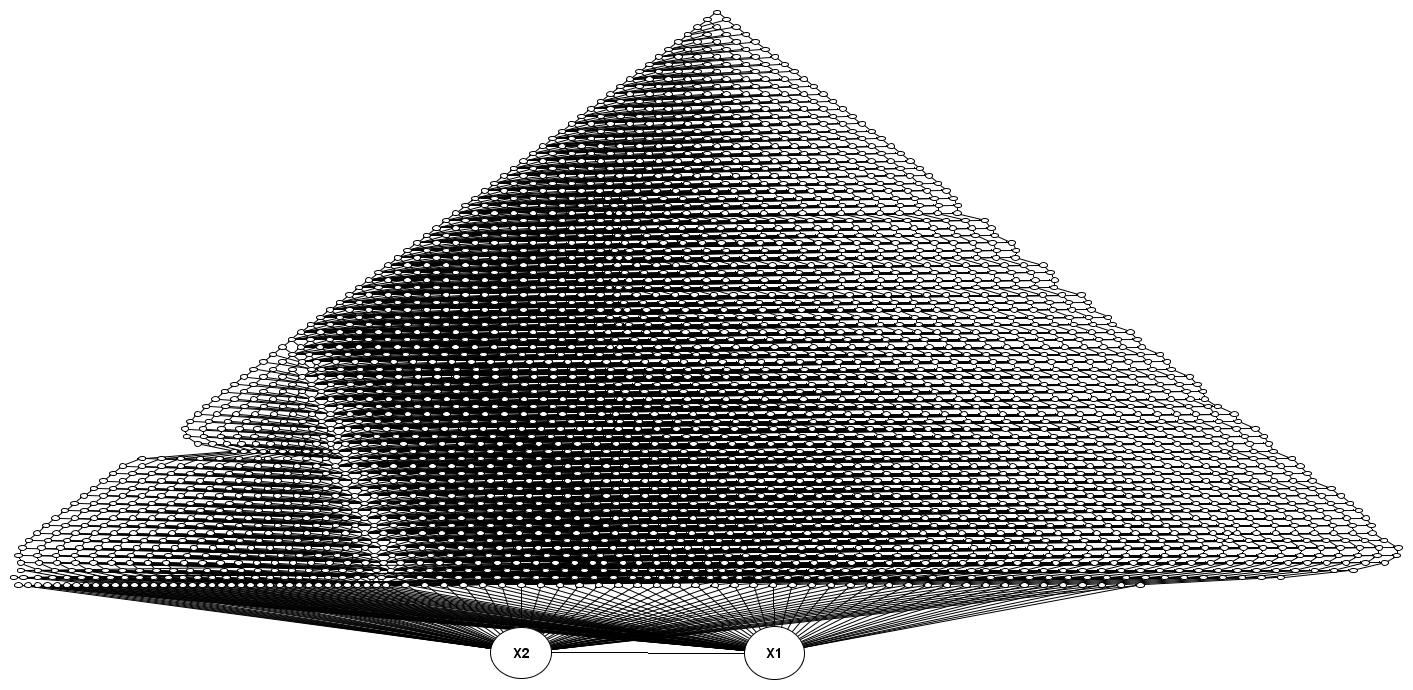}
 	\centering
 	\caption{$\kappa_{80}$ BFG model with node $X_1$ and $X_2$ presented }
 \end{figure}
\\Figure 13 illustrates the $\kappa_{80}$ BFG model we presented in the paper (there are 3083 variables in this BFG). Obviously there are large connections from one node to others, such as node $X_1$ and $X_2$, which will result in the same node appearing in many different regions in a region graph and introducing numerical unstable problems. So if the first 20 dimensions are accurately approximated, higher dimension variables must be also accurate by the approximation. If otherwise the inaccuracy will appear in low dimensions (i.e. the first 20 dimensions) in priority.
\ifCLASSOPTIONcompsoc
  \section*{Acknowledgments}
\else
  \section*{Acknowledgment}
\fi
This work is supported by European Research Council Advanced Grant. The full ERC code is ERC-2013-AdG339182-BAYES-KNOWLEDGE.

\ifCLASSOPTIONcaptionsoff
  \newpage
\fi



%

{

	\bibliographystyle{ieeetr}
	\bibliography{IEEEexample}
}

%

\begin{IEEEbiographynophoto}{Peng Lin}
is PostDoc at School of EECS, Queen Mary U. of London.
\end{IEEEbiographynophoto}

\begin{IEEEbiographynophoto}{Martin Neil}
is Professor of statistics at School of EECS, Queen Mary U. of London.
\end{IEEEbiographynophoto}


\begin{IEEEbiographynophoto}{Norman Fenton}
is Professor of Computer Science at School of EECS, Queen Mary U. of London.
\end{IEEEbiographynophoto}




\end{document}